\documentclass[journal]{IEEEtran}
\IEEEoverridecommandlockouts

\usepackage{cite}
\newtheorem{theorem}{\bf Theorem}
\newtheorem{lemma}{\bf Lemma}

\usepackage{mathrsfs}
\usepackage{lipsum}
\usepackage{graphicx}
\usepackage{float}
\usepackage{amsfonts}
\usepackage{dsfont}
\usepackage{setspace}
\usepackage[ruled,linesnumbered,noend]{algorithm2e}
\usepackage{algpseudocode}
\usepackage{threeparttable}

\usepackage{array}

\usepackage[utf8]{inputenc} 
\usepackage[T1]{fontenc}    
\usepackage{graphicx} \graphicspath{ {figures/} }
\usepackage{times}
\usepackage{graphicx}
\usepackage{amsmath,amssymb}
\usepackage{acronym}
\usepackage{enumitem}
\usepackage[breaklinks,colorlinks,linkcolor=black]{hyperref}
\usepackage{balance}
\usepackage{xspace}
\usepackage{setspace}
\usepackage[skip=3pt,font=small]{subcaption}
\usepackage[skip=3pt,font=small]{caption}
\usepackage[dvipsnames,svgnames,x11names]{xcolor}
\usepackage[capitalise,nameinlink]{cleveref}
\usepackage{booktabs,tabularx,colortbl,multirow,multicol,array,makecell}
\usepackage{dblfloatfix}
\usepackage[misc]{ifsym}
\usepackage{pifont}
\usepackage{cite}

\makeatletter
\DeclareRobustCommand\onedot{\futurelet\@let@token\@onedot}
\def\@onedot{\ifx\@let@token.\else.\null\fi\xspace}

\makeatother

\makeatletter
\def\BState{\State\hskip-\ALG@thistlm}
\makeatother

\makeatletter
\renewcommand{\paragraph}{%
  \@startsection{paragraph}{4}%
  {\z@}{0ex \@plus 0ex \@minus 0ex}{-1em}%
  {\hskip\parindent\normalfont\normalsize\bfseries}%
}
\makeatother

\crefname{algorithm}{Alg.}{Algs.}
\Crefname{algocf}{Algorithm}{Algorithms}
\crefname{section}{Sec.}{Secs.}
\Crefname{section}{Section}{Sections}
\crefname{table}{Tab.}{Tabs.}
\Crefname{table}{Table}{Tables}
\crefname{figure}{Fig.}{Fig.}

\definecolor{gblue}{HTML}{4285F4}
\definecolor{gred}{HTML}{DB4437}
\definecolor{ggreen}{HTML}{0F9D58}

\definecolor{mygray}{gray}{.92}

\begin{document}

\title{ReSPIRe: Informative and Reusable Belief Tree Search for Robot Probabilistic Search and Tracking in Unknown Environments}

\author{Kangjie Zhou, 
Zhaoyang Li,
Han Gao,
Yao Su,~\IEEEmembership{Member, IEEE},
Hangxin Liu,~\IEEEmembership{Member, IEEE},\\
Junzhi Yu,~\IEEEmembership{Fellow, IEEE}, and Chang Liu,~\IEEEmembership{Member, IEEE}
\thanks{This work is sponsored by Beijing Nova Program (20220484056, 20240484498), National Natural Science Foundation of China (62203018), and State Key Laboratory of Intelligent Green Vehicle and Mobility (KFZ2410) (\textit{Corresponding author: Chang Liu.})}
\thanks{Kangjie Zhou, Han Gao, Junzhi Yu, and Chang Liu are with School of Advanced Manufacturing and Robotics, Peking University, Beijing 100871, China (emails: kangjiezhou@pku.edu.cn; hangaocoe@pku.edu.cn; junzhi.yu@ia.ac.cn; changliucoe@pku.edu.cn).}
\thanks{Zhaoyang Li is with Institute of Automation, Chinese Academy of Sciences, Beijing, China (email: lizhaoyang2025@ia.ac.cn).}
\thanks{Yao Su and Hangxin Liu are with State Key Laboratory of General Artificial Intelligence, Beijing Institute for General Artificial Intelligence (BIGAI), Beijing 100080, China (emails: suyao@bigai.ai; liuhx@bigai.ai).} 
}
\maketitle

\begin{abstract}

Target search and tracking (SAT) is a fundamental problem for various robotic applications such as search and rescue and environmental exploration. 
This paper proposes an informative trajectory planning approach, namely ReSPIRe, for SAT in unknown cluttered environments under considerably inaccurate prior target information and limited sensing field of view. 
We first develop a novel sigma point-based approximation approach to fast and accurately estimate mutual information reward under non-Gaussian belief distributions, utilizing informative sampling in state and observation spaces to mitigate the computational intractability of integral calculation. 
To tackle significant uncertainty associated with inadequate prior target information, we propose the hierarchical particle structure in ReSPIRe, which not only extracts critical particles for global route guidance, but also adjusts the particle number adaptively for planning efficiency. 
Building upon the hierarchical structure, we develop the reusable belief tree search approach to build a policy tree for online trajectory planning under uncertainty, which reuses rollout evaluation to improve planning efficiency. 
Extensive simulations and real-world experiments demonstrate that ReSPIRe outperforms representative benchmark methods with smaller MI approximation error, higher search efficiency, and more stable tracking performance, while maintaining outstanding computational efficiency.
\end{abstract}

\begin{IEEEkeywords}
Search and tracking, information gathering, mutual information, Monte Carlo tree search.
\end{IEEEkeywords}

\section{Introduction}
\IEEEPARstart{T}{arget} search and tracking (SAT) using autonomous robots plays a pivotal role in numerous applications such as surveillance~\cite{lozano2022surveillance,papaioannou2023distributed}, disaster response~\cite{senthilnath2024metacognitive, cui2025cooperative}, and environment exploration~\cite{wang2023active, yu2025cognitive}. 
In these applications, the robot first explores the environment and searches for the lost target. 
Once the target is detected, the robot enters the tracking stage to maintain the target inside the field of view (FOV).
Facing intrinsic uncertainties in the realistic world such as imprecise prior target information and sensor noise, previous works~\cite{ tisdale2009autonomous,furukawa2006recursive,cui2015mutual} have utilized filtering approaches to estimate the target \textit{belief}, the probability distribution of the target state, to account for target state uncertainty.

Building upon the belief estimation, mainstream strategies formulate SAT as an information-gathering problem, which involves active control of the robot to acquire informative measurements to reduce the uncertainty~\cite{charrow2014approximate, hoffmann2009mobile,legrand2022cell}.

\begin{figure}[!t] 
\centering
\includegraphics[width=\linewidth]{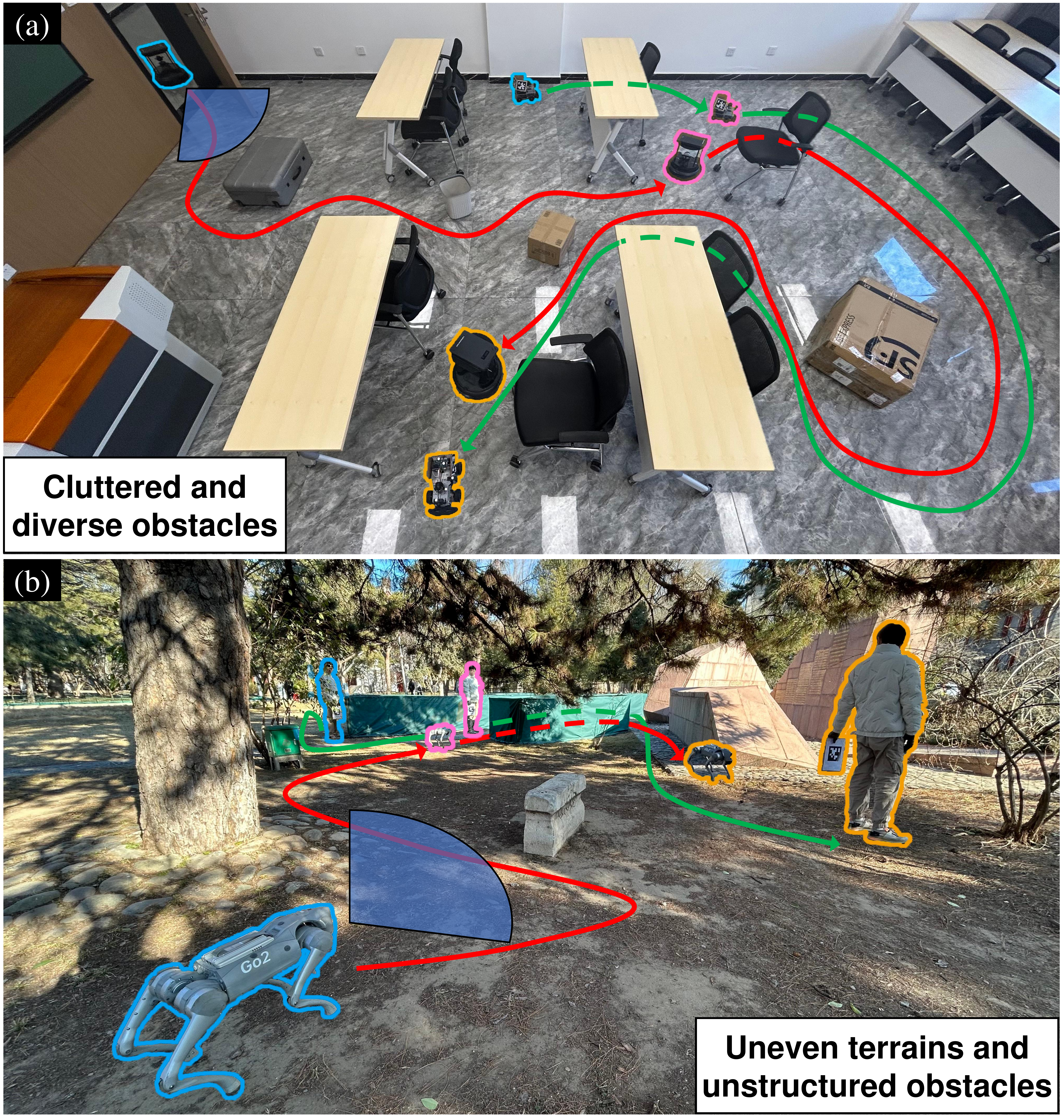} 
\caption{\textbf{The process of target SAT in unknown (a) indoor cluttered environment and (b) outdoor environment with irregular terrains and unstructured obstacles.}
The robot and target trajectories are displayed in red and green, respectively.
The contour represents different stages: cyan contour for target search, pink contour for target being found, and orange contour for target tracking.
}
\label{Fig:cover_picture} 
\end{figure}

Various planning methods have been utilized to tackle information gathering, including greedy strategies~\cite{julian2012distributed,legrand2022cell}
, sampling-based methods~\cite{ghaffari2019sampling,schmid2020efficient,wang2019semantic}, and optimization-based methods~\cite{chen2023flying,tao2021path}.
In contrast to these traditional techniques, the partially observable Markov decision process (POMDP) offers a rigorous mathematical framework for planning under uncertainty. 
When incorporated with an information-theoretic reward function, POMDP offers a potential solution to the information-gathering problem~\cite{best2019dec,fischer2020information}. 
However, both solving POMDP and calculating information-theoretic reward introduce significant computational burdens, hindering the online planning capability.
Moreover, most works in SAT concentrate on obstacle-free or known environments and rely on a fairly accurate prior knowledge of the target.
However, such assumptions fall short in real-world scenarios like post-disaster ruins, which are characterized by unknown complex obstacles and lack of prior target information, rendering existing methods inapplicable to practical situations.

To overcome these challenges, 
we propose the \underline{re}usable belief tree search with \underline{s}igma \underline{p}oint-based mutual \underline{i}nformation \underline{re}ward approximation (ReSPIRe), 
a computationally efficient informative planner, 
to generate non-myopic trajectories for mobile target SAT in complex unknown environments (\Cref{Fig:cover_picture}).
The proposed approach is able to react swiftly in the presence of considerable target uncertainties and cluttered obstacles, allowing for real-time and safe planning to accomplish SAT tasks in complex unknown environments.
The main contributions can be summarized as follows:
\begin{itemize}
\item We propose a novel sigma point (SP)-based approximation approach to computing the predictive mutual information (MI) under continuous state and observation spaces for 
non-parametric belief states, while taking the limited sensing FOV into account.
A theoretical analysis of the approximation error is also provided.

\item We present a hierarchical particle structure that extracts refined information from dispersed particles to balance global route efficiency and local uncertainty reduction, while adaptively adjusting particle number for efficiency.

\item We develop the reusable belief tree search (RBTS), a tree-based planner that incorporates a recycling step into Monte Carlo Tree Search (MCTS), to reuse the rollout evaluation for planning efficiency.
\end{itemize}
We have conducted extensive simulations and real-world experiments for validation, and the results show that the proposed method can accomplish SAT tasks in unknown environments with superior search efficiency, enhanced estimation accuracy, and real-time computational capability.

This article significantly extends our previous conference paper~\cite{zhou2024aspire} with the following aspects. 
First, this work investigates SAT tasks within a more intricate scenario including unknown obstacles and substantial prior misinformation, which is a practical yet less studied problem.
Second, we provide a theoretical analysis to derive the approximation error bound of the MI approximation method presented in~\cite{zhou2024aspire}.
Third, we propose a hierarchical particle representation and a reusable tree-based planner for computationally efficient trajectory generation under significant uncertainty. 
Last, we evaluate the proposed methods comprehensively with new simulations and real-world experiments.

The organization of this paper is as follows. 
Section II introduces related work. 
Section III presents the background knowledge and problem formulation.
Section IV proposes the informative planner, ReSPIRe. 
Simulation and experimental results are then provided in Sections V and VI, respectively.

\section{Related Work}
\subsection{Probabilistic Search and Tracking}
Early works of probabilistic SAT exploited grid-based Bayesian filtering for target state estimation and adopted probability of detection (POD) as the objective function to guide the robot decision-making~\cite{furukawa2006recursive,tisdale2009autonomous,xiao2017sampling}.
However, the localization accuracy is limited by the discretization resolution, and the POD metric only measures the probability of finding the target, lacking the ability to quantify the target uncertainty.
Recently, enormous studies have shown that information-theoretic objectives, especially the mutual information~\cite{charrow2014approximate,schlotfeldt2018anytime,zhou2024aspire}, demonstrate superior performance in encouraging robots to proactively gather target information and decrease the target state estimation uncertainty.
However, these works assume obstacle-free or known environments, hindering their applicability to real-world situations.
To adapt to unknown environments, Wolek et al.~\cite{wolek2020cooperative} presented an information-theoretic method for simultaneous mapping and target search.
Nevertheless, the target state space is discretized for simplification, and obstacle avoidance is not considered.

\subsection{Motion Planning for Information Gathering}
Diverse planning approaches have been utilized for information gathering.
The greedy policy that chooses the next-best-view (NBV) has been widely adopted for its low computational complexity~\cite{julian2012distributed,legrand2022cell}, yet such a strategy usually falls short in complex environments due to its myopic nature. 
In response to this challenge, 
Jadidi et al.~\cite{ghaffari2019sampling} proposed incremental sampling-based methods for non-myopic informative path planning, which enables online replanning by incorporating a criterion for automatic algorithm termination. 
Nevertheless, the sampling strategy and amount are challenging to determine to reach a reasonable trade-off between optimality and efficiency.
Liu et al.~\cite{liu2017model} used model predictive control to generate informative trajectories for SAT in continuous spaces, but the restrictive assumption of the Gaussian belief state limits the generality of the method.

The POMDP is a principled mathematical framework for modeling sequential decision-making tasks under uncertainty, while solving POMDPs exactly is computationally intractable due to the curse of dimensionality and history. 
To reduce the computational complexity, abundant studies have attempted to obtain approximate solutions for POMDP~\cite{wandzel2019multi,xiao2019online}.
For instance, Xiao et al.~\cite{xiao2019online} leveraged a tree-based POMDP solver for target search, which adjusts the action space adaptively based on the current belief.
However, existing methods sample states from the belief and only update the state, hindering their applicability from information-gathering problems where the reward is a belief-dependent function.

\subsection{MCTS-Based Planning Method}

MCTS has gained popularity recently as an effective approach for non-myopic planning due to its ability to allocate computation resources to more valuable subtrees to prevent exhaustive search, and has been utilized in trajectory planning for information gathering and SAT tasks~\cite{ goldhoorn2018searching, best2019dec}. 
Albeit MCTS was initially utilized to solve observable problems, Silver et al.~\cite{silver2010monte} extend MCTS to partially observable settings and obtained a tree-based solver for POMDP in large domains, which uses Monte Carlo sampling to reduce the inherent computational burden.
Following this insight, multiple tree-based POMDP solvers are presented~\cite{somani2013despot,kurniawati2016online}.
However, since the sample-based methods only update state rather than belief, these methods are unable to accommodate belief-dependent rewards.
To overcome this gap, the particle filter tree (PFT) is proposed, which leverages MCTS for tree growth and the particle filter for belief update~\cite{sunberg2018online,fischer2020information}.
However, since the belief update and belief-dependent reward computation are both computationally expensive, these methods only employ a limited number of particles to avoid the enormous computational burden, which compromises estimation accuracy and therefore cannot meet the needs of many realistic tasks with significant target uncertainty.

\subsection{Mutual Information Approximation}
MI is a widely adopted information-theoretic metric in information-gathering problems, as maximizing MI between belief states and predicted measurements drives robots to obtain informative observations to decrease uncertainty.
However, calculating MI for a non-Gaussian belief involves integration over the continuous state and observation spaces, and lacks a general analytical expression. 
Monte Carlo integration could be used, but may incur high computational expenses, hindering its use for online planning.  
To alleviate the computational burden, space discretization is utilized to sacrifice state estimation accuracy in favor of computational simplicity~\cite{zhang2020fsmi,asgharivaskasi2023semantic}.
Other works employed the particle filter to sample the continuous state space, facilitating MI computation~\cite{ryan2010particle, charrow2014approximate}. 
However, integrating over the continuous observation space remains a significant hurdle. 
To mitigate this issue, several approximation approaches are proposed for analytical integration calculation~\cite{huber2008entropy,kolchinsky2017estimating}.
Nevertheless, the approximation accuracy is compromised in favor of computational efficiency, leading to non-trivial approximation error.

\section{Problem Formulation}
We consider a robot searching and tracking a mobile target in an unknown cluttered environment with limited sensing FOV and significant initial uncertainty of target state.
We first introduce system models and formulate the SAT problem.
\subsection{System Models \label{subsec:system-model}}
\subsubsection{Motion Model}

Consider a discrete-time robot dynamics
\begin{equation}
\begin{split}
\boldsymbol{x}_{k+1}^{r}=\mathbf{f}^r(\boldsymbol{x}_{k}^{r},\boldsymbol{u}_{k}^{r}),
\end{split}
\label{eqn:1}
\end{equation}
where $\boldsymbol{x}_{k}^{r}$ and $\boldsymbol{u}_{k}^{r}$ denote the robot state and control, respectively, and the robot dynamics function $\mathbf{f}^r$ is assumed known.

Denote the target state as $\boldsymbol{x}_{k}^{t}$, and the target moves with a stochastic motion model
\begin{equation}
\begin{split}
\boldsymbol{x}_{k+1}^{t}=\mathbf{f}^t (\boldsymbol{x}_{k}^{t})+\boldsymbol{\eta}_{k},\boldsymbol{\eta}_{k}\sim\mathcal{N}(0,\mathbf{Q}),
\end{split}
\label{eqn:motion model}
\end{equation}
with the dynamics function $\mathbf{f}^t$ unknown.
Here $\boldsymbol{\eta}_k$ is a zero-mean Gaussian process noise with covariance matrix $\mathbf{Q}$.

\subsubsection{Observation Model}
Due to the limited sensing domain and obstacle occlusion, no target measurement can be obtained when the target is outside the FOV. 
To reflect the intermittency of target measurements, a binary parameter $\gamma_{k}$ is defined to indicate if the target is inside the FOV ($\gamma_{k}=1$) or not ($\gamma_{k}=0$), and the target observation model is
\begin{equation}
\begin{split}
\boldsymbol{z}_{k}=\begin{cases}
\mathbf{h}(\boldsymbol{x}_{k}^{r},\boldsymbol{x}_{k}^{t})+\boldsymbol{\varepsilon}_{k},\boldsymbol{\varepsilon}_{k}\sim\mathcal{N}(0,\mathbf{\Sigma}) & \;\gamma_{k}=1\\
\varnothing & \;\gamma_{k}=0
\end{cases},
\end{split}
\label{eqn:sensing model}
\end{equation}
where $\boldsymbol{z}_{k} \in \mathbb{R}^m \cup \varnothing$ denotes the target measurement and $\boldsymbol{\varepsilon}_{k}$
is a Gaussian white noise with covariance matrix $\mathbf{\Sigma}$.

\subsection{Target Belief Estimation using Particle Filter}
To account for potential nonlinearity in target dynamics and sensor models, especially due to the limited sensing domain, we use the particle filter to estimate the target belief because of its capability to represent arbitrary probability distributions.
Denote $\boldsymbol{b}_k = P(\boldsymbol{x}_{k}^t|\boldsymbol{u}_{0:k-1}^t,\boldsymbol{z}_{1:k}) $ as the target belief, and $\boldsymbol{b}_k$ can be approximated by weighted particles as
\begin{equation}
\begin{split}
\boldsymbol{b}_k \approx\sum\nolimits_{j=1}^{N}w_{k}^{j}\delta(\boldsymbol{x}_{k}^{t}-\tilde{\boldsymbol{x}}_{k}^{t,j}),
\end{split}
\end{equation} 
where $\tilde{\boldsymbol{x}}_{k}^{t,j}$ is the $j$th particle, $w_{k}^{j}$ is the corresponding weight, $N$ is the number of particles, and $\delta(\cdot)$ is a Dirac function. 
The particle filtering processes with the following prediction step (\cref{eqn:pf_pred}) and update step (\cref{eqn:pf_upd}) to obtain $\boldsymbol{b}_{k+1}$,

\noindent \textbf{Prediction}. Particle states are forward predicted based on the target dynamics,
\begin{equation}\label{eqn:pf_pred}
\begin{split}
\tilde{\boldsymbol{x}}_{k+1}^{t,j}\sim\mathcal{N}(\mathbf{f}^t(\tilde{\boldsymbol{x}}_{k}^{t,j}),\mathbf{Q}),\quad j=1,\ldots,N,
\end{split}
\end{equation}
\textbf{Update}. Particle weights are updated with new measurements,
\begin{equation}\label{eqn:pf_upd}
\begin{split}
w_{k+1}^{j}=\frac{P(\boldsymbol{z}_{k+1}|\tilde{\boldsymbol{x}}_{k+1}^{t,j})}{\sum_{j=1}^{N}P(\boldsymbol{z}_{k+1}|\tilde{\boldsymbol{x}}_{k+1}^{t,j})} w_{k}^{j},\quad j=1,\ldots,N,
\end{split}
\end{equation}
To alleviate particle degeneracy, we subsequently perform low variance resampling strategy to mitigate sampling error.

\subsection{Problem Formulation of SAT}
We formulate the SAT problem as a finite-horizon belief Markov decision process (MDP) defined by a tuple $(\mathcal{B},\mathcal{A},\tau,R,h,\gamma)$, with belief state space $\mathcal{B}$, action space $\mathcal{A}$, belief transition model $\tau$, planning horizon $h$, discount factor $\gamma$, and reward function $R$.
The belief state is defined as $\boldsymbol{B}_k = [\boldsymbol{x}_{k}^r, \boldsymbol{b}_{k}]\in \mathcal{B}$, where the robot state $\boldsymbol{x}_{k}^r$ is assumed to be fully observable.
The action $\boldsymbol{a}_k = \boldsymbol{u}_k^{r} \in \mathcal{A}$ denotes the control inputs of the robot.
The belief transition model $\tau$ is defined as $\boldsymbol{B}_{k+1}=[\boldsymbol{x}_{k+1}^r, \boldsymbol{b}_{k+1}]=\tau(\boldsymbol{B}_k,\boldsymbol{a}_k,\boldsymbol{z}_k)$, where $\boldsymbol{x}_{k+1}^r$ is propagated based on \cref{eqn:1} and $\boldsymbol{b}_{k+1}$ is updated with the particle filtering.
The initial target belief $\boldsymbol{b}_{0}$ is configured with broadly distributed particles, indicative of substantial uncertainty.
The objective is to obtain the optimal action sequence $\boldsymbol{a}_{k:k+h-1}^{\ast}$ that maximizes the expected total discounted reward, 
\begin{equation}
\begin{split}
\boldsymbol{a}_{k:k+h-1}^{\ast}=\mathop{\arg\max}\limits_{\boldsymbol{a}_{k:k+h-1}} \mathbb{E} \left[ \sum\nolimits_{t=k}^{k+h-1} \gamma^{t-k}R(\boldsymbol{B}_{t},\boldsymbol{a}_{t}) \right],
\nonumber
\end{split}
\end{equation}
where $\mathbb{E}$ is the expectation over future beliefs. 
To encourage the robot to gather more target information from future observations, we define the reward $R$ as the MI between the target belief and predicted measurements,
\begin{equation}
\begin{split}
R(\boldsymbol{B}_{k},\boldsymbol{a}_{k}) &= I(\boldsymbol{x}_{k+1}^t;\boldsymbol{z}_{k+1}),
\end{split}
\end{equation}
where $I$ denotes the MI.
The robot then executes the first optimal action $\boldsymbol{a}_{k}^{\ast}$ and replans at the next time step based on the new measurements in a receding-horizon manner. 

\section{ReSPIRe Trajectory Planner}
\begin{figure}[!t] 
\centering
\includegraphics[width=0.49\textwidth]{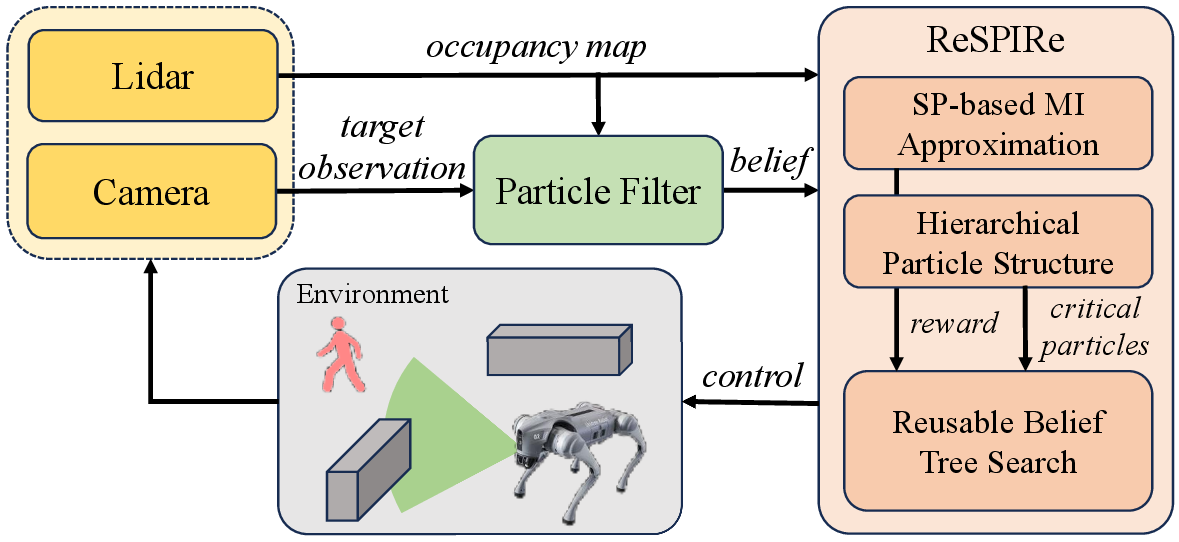} 
\caption{\textbf{The overview of the proposed framework for SAT under unknown environments.}} 
\label{Fig:overview} 
\end{figure}

This section proposes the ReSPIRe, an uncertainty-aware online planning algorithm to generate informative trajectories for SAT in unknown cluttered environments. 
\cref{Fig:overview} presents the overall procedure of the proposed SAT framework.
As the robot navigates in an unknown environment, onboard sensors obtain the observations for target estimation (camera) and mapping (Lidar).
Given the estimated belief after particle filtering and current occupancy map $\mathcal{M}$, ReSPIRe executes as a trajectory planner and outputs the optimal control.
ReSPIRe consists of three main components: First, a sigma-point (SP) based MI approximation approach is presented for accurate and efficient reward calculation (\cref{subsec:mutual-information}).
Second, a particle hierarchy approach is proposed with adaptive space partition to extract refined estimation information for long-term decision guidance and fast computation of the reward function (\cref{subsec:particle-hierarchy}).
Last, the reusable belief tree search is developed to construct an asymmetric belief tree to generate informative trajectories while avoiding collision in real time (\cref{subsec:RCTS}). 

It is noteworthy that ReSPIRe incorporates two critical components for computational efficiency enhancement to enable online planning under complex belief space.
First, the hierarchical particle representation not only extracts abstract information from particles for overall planning efficiency, but also adaptively adjusts the number of particles to enable fast MI computation and tree construction.
Moreover, we develop a novel recycling procedure in the standard tree search algorithm, which leverages previous rollout evaluation to estimate the future accumulated reward, and creates multiple new nodes in parallel in one iteration, thereby improving the algorithm efficiency of the proposed algorithm.
These properties enable the retention of abundant particles to ensure accurate belief estimation, and simultaneously promise a desirable trade-off between the effectiveness and efficiency of the planning stage.

\subsection{SP-Based MI Approximation \label{subsec:mutual-information}}

\subsubsection{Reward Function Definition}
According to the properties of MI, we can derive that
\begin{equation}
\begin{split}
I(\boldsymbol{x}_{k+1}^t;\boldsymbol{z}_{k+1}) = H(\boldsymbol{z}_{k+1})-H(\boldsymbol{z}_{k+1}|\boldsymbol{x}_{k+1}^t).
\end{split}
\label{obj_def}
\end{equation}
We can approximate the future belief with the particle representation,
\begin{equation}
\begin{split}
P(\boldsymbol{x}_{k+1}^t) \approx \sum\nolimits_{j=1}^{N}w_{k}^{j}\delta(\boldsymbol{x}_{k+1}^t-\tilde{\boldsymbol{x}}_{k+1}^{t,j}),
\end{split}
\label{eqn:10}
\end{equation}
where the particle state is predicted based on the target dynamics, while the weight remains unchanged since the future observation is unknown. 
Denote $S_f$ as the set of indices of particles inside the FOV.
Based on the observation model and particle expression, we can obtain
\begin{equation}
\resizebox{0.89 \linewidth}{!}{%
$\begin{aligned}
P(\boldsymbol{z}_{k+1}|\tilde{\boldsymbol{x}}_{k+1}^{t,j}) 
=  \begin{cases}
\mathcal{N}(\boldsymbol{z}_{k+1};\mathbf{h}(\boldsymbol{x}_{k+1}^{r},\tilde{\boldsymbol{x}}_{k+1}^{t,j}),\mathbf{\Sigma}) & j \in S_f\\
{\mathds{1}}_{\boldsymbol{z}_{k+1}=\varnothing} & j \notin S_f
\end{cases},
\end{aligned}$}
\label{eqn:p1}
\end{equation}
and
\begin{equation}
\begin{split}
P(\boldsymbol{z}_{k+1}) & = \int P(\boldsymbol{z}_{k+1}|\boldsymbol{x}_{k+1}^t)P(\boldsymbol{x}_{k+1}^t)d\boldsymbol{x}_{k+1}^t \\
& \approx \sum\nolimits_{j=1}^{N}w_{k}^{j}P(\boldsymbol{z}_{k+1}|\tilde{\boldsymbol{x}}_{k+1}^{t,j}).
\end{split}
\label{eqn:p2}
\end{equation}
According to the observation model, the variable $\boldsymbol{z}_{k+1}$ is a continuous-discrete mixed variable defined in $\mathbb{R}^m \cup \varnothing$.
Based on the definition of entropy for mixed random variable~\cite{nair2006entropy}, we can derive following theorem.
\begin{theorem}[Entropy Computation under Particle Expression and Limited FOV]
\label{thm:definition}
With the particle expression (\cref{eqn:10}) and probability density function (\cref{eqn:p1} and \cref{eqn:p2}), the entropy in \cref{obj_def} can be calculated as
\begin{equation}
\resizebox{\linewidth}{!}{%
$\begin{aligned}
H(\boldsymbol{z}_{k+1}|\boldsymbol{x}_{k+1}^t) = \sum\nolimits_{j=1}^{N}w_{k}^{j} {\mathds{1}}_{j\in S_f} \left[\frac{m}{2}(\log2\pi+1)+\frac{1}{2}\log|\mathbf{\Sigma}|\right].
\nonumber
\end{aligned}$}
\end{equation}
\begin{equation}
\begin{split}
H(\boldsymbol{z}_{k+1})= -p_{\varnothing}\log p_{\varnothing} + H_r.
\nonumber
\end{split}
\end{equation}
where 
\begin{subequations}
\begin{equation}
p_{\varnothing} = P(\boldsymbol{z}_{k+1} = \varnothing) = \sum\nolimits_{j\notin S_f} w_{k}^{j},
\end{equation}
\begin{equation}
H_r = - \int_{\mathbb{R}^m} p_{r}\log p_{r}d\boldsymbol{z}_{k+1},
\end{equation}
\begin{equation}
p_{r} = \sum\nolimits_{j\in S_f} w_{k}^{j} P(\boldsymbol{z}_{k+1}|\tilde{\boldsymbol{x}}_{k+1}^{t,j}).
\end{equation}
\end{subequations}
\end{theorem} 

The proof is provided in the Appendix A\footnote{Please refer to the appendix in the supplementary material.}.
Note that $p_{r}$ follows a Gaussian Mixture Model (GMM), whose entropy has no analytical expression. 
In this regard, we propose to utilize the sigma points associated with each Gaussian component in GMM to approximate the entropy $H_r$, which will be detailed in the next subsection.

\subsubsection{SP-Based Entropy Approximation}

The key idea behind the SP-based approximation is utilizing the property of sigma points~\cite{van2004sigma} that keep the first two moments of the distribution invariant to estimate the Gaussian component in GMM and obtain an analytical expression of the GMM entropy.

Denote the sigma points and their weights associated with the $j$th Gaussian component as $[\tilde{\boldsymbol{z}}_{k+1}^{j,0},\ldots,\tilde{\boldsymbol{z}}_{k+1}^{j,2m}]^{T}$ and $\left[w_{s}^{j,0}, \ldots,w_{s}^{j,2m}\right]^{T},\forall j \in S_f$, respectively, and they are computed as follows~\cite{van2004sigma},
\begin{equation}
\resizebox{0.89\linewidth}{!}{%
$\begin{aligned}
\tilde{\boldsymbol{z}}_{k+1}^{j,0} & =  \boldsymbol{\mu}_{j},\\
\tilde{\boldsymbol{z}}_{k+1}^{j,l} & =  \boldsymbol{\mu}_{j}+\big(\sqrt{(\lambda+m)\mathbf{\Sigma}}\big)_{l}, \quad l=1,\ldots,m\\
\tilde{\boldsymbol{z}}_{k+1}^{j,l} & =  \boldsymbol{\mu}_{j}-\big(\sqrt{(\lambda+m)\mathbf{\Sigma}}\big)_{l-m}, \quad l=m+1,\ldots,2m\\
w_{s}^{j,0} & = \frac{\lambda}{\lambda+m},\;w_{s}^{j,l} =  \frac{1}{2(\lambda+m)}, \quad l=1,\ldots,2m
\end{aligned}$}
\label{eqn:sigma_points}
\end{equation}
where $\lambda$ is a parameter that determines the sigma points spread, $\boldsymbol{\mu}_{j}=\mathbf{h}(\boldsymbol{x}_{k+1}^{r},\tilde{\boldsymbol{x}}_{k+1}^{t,j})$ is the mean of $P(\boldsymbol{z}_{k+1}|\tilde{\boldsymbol{x}}_{k+1}^{t,j})$, and $\big(\sqrt{(\lambda+m)\mathbf{\Sigma}}\big)_{l}$ is the $l$-th column of the matrix square root. 
Since the observation is $m$-d, there are $2m+1$ sigma points for each Gaussian component.
Based on sigma points, we approximate $H_r$ with the following theorem.
\begin{theorem}[SP-based GMM Entropy Approximation]
\label{thm:sp}
With sigma points in \cref{eqn:sigma_points}, $H_r$ can be approximated by
\begin{equation} \label{eqn:H_r}
\resizebox{0.89\linewidth}{!}{%
$\hat{H_r}  = - \sum\nolimits_{j\in S_f} w_{k}^{j}\sum\nolimits_{l=0}^{2m}w_{s}^{j,l}\log\sum\nolimits_{i\in S_f}  w_{k}^{i}P(\tilde{\boldsymbol{z}}_{k+1}^{j,l}|\tilde{\boldsymbol{x}}_{k+1}^{t,i}).$}
\end{equation}
Furthermore, there exists a small positive constant $c$ such that the approximation error is bounded by
\begin{equation}
\begin{split}
|H_r-\hat{H_r}| \leq c m\sigma_{\max}^2.
\end{split}
\end{equation}
where $m$ is the observation dimension and $\sigma_{\max}^2$ is the maximum eigenvalue of $\mathbf{\Sigma}$ in \cref{eqn:sigma_points}.
\end{theorem}

The proof is provided in the Appendix B.
Utilizing sigma points, we obtain an explicit expression to approximate the entropy $H_r$, and the approximation error of the SP-based method is bounded theoretically by the observation dimension and the sensor noise.
Moreover, the simulation results in \cref{subsec:MI_comparison} also demonstrate the approximation accuracy of the proposed method under different sensor noise.

\subsubsection{Simplification and Truncation of Particles}
Note that the computational complexity of \cref{eqn:H_r} is quadratic with respect to the number of particles, which hinders the online planning process when encountering numerous particles. 
We adopt two approaches for efficient implementations of the proposed approximation method.
First, we apply the particle simplification technique similar to~\cite{charrow2014approximate}, which partitions the state space
and generates a simplified particle set by replacing particles
in the same cell with their weighted average.
From the theoretical proof in~\cite{charrow2014approximate}, the approximation error of particle simplification is negligible when the grid size is smaller than the observation model noise. 
We can modify the length of partitioned grids to reach a promising balance between computational efficiency and approximation error.
Second, notice that when the distance between $i^{th}$ and $j^{th}$ particle is large, the Gaussian probability $P(\tilde{\boldsymbol{z}}_{k+1}^{j,l}|\tilde{\boldsymbol{x}}_{k+1}^{t,i})$ in \cref{eqn:H_r} becomes sufficiently as the variable deviates from the mean, thereby having a negligible impact on the MI result.
To account for this, we define a particle set $S_a^j$, which contains particles within specified truncation range of the $j$th particle, and approximate \cref{eqn:H_r} as follows:
\begin{equation} \label{eqn:sp_truc}
\begin{split}
H_r \approx - \sum\nolimits_{j\in S_f}\sum\nolimits_{l=0}^{2m}w_{s}^{j,l}\log\sum\nolimits_{i\in S_f\cap S_a^j}  w_{k}^{i}P(\tilde{\boldsymbol{z}}_{k+1}^{j,l}|\tilde{\boldsymbol{x}}_{k+1}^{t,i}).
\nonumber
\end{split}
\end{equation}
Considering only the contributions of adjacent particles, this refinement improves computational efficiency with negligible loss of approximation accuracy.

\subsection{Hierarchical Particle Representation \label{subsec:particle-hierarchy}}
\begin{figure}[!t] 
\centering
\includegraphics[width=0.49\textwidth]{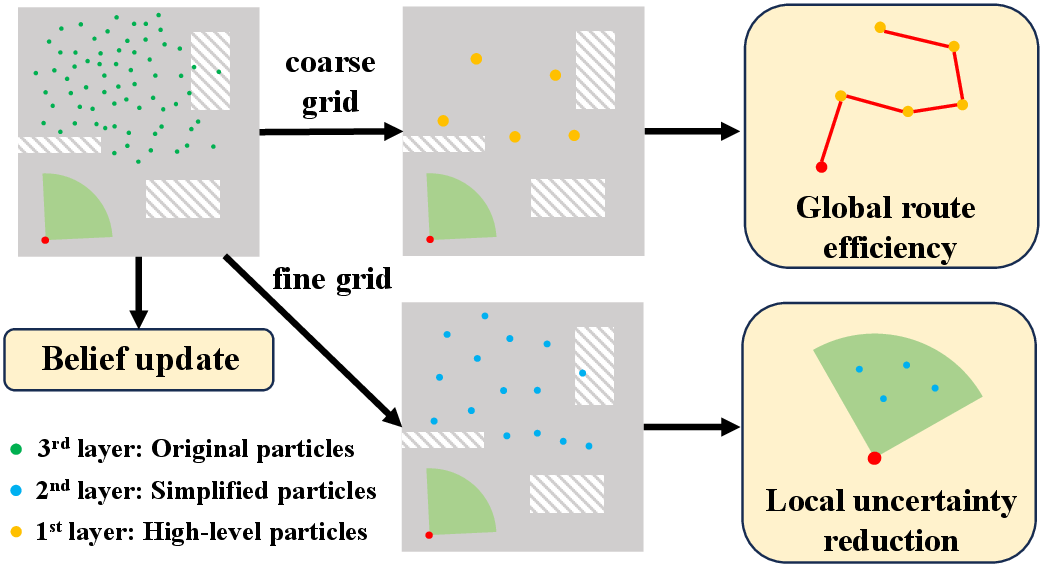} 
\caption{\textbf{Illustration of the three-layer hierarchical particle structure.}
Along with the original particles for belief update, refined particles that cluster by grids of different sizes are used for overall route efficiency and local uncertainty reduction.
} 
\label{Fig:particle_hierarchy} 
\end{figure}

To improve planning performance under considerable uncertainty characterized by dispersed particles, a three-layer hierarchical particle structure is devised to extract refined information from original particles while adjusting the particle number adaptively.
As depicted in \Cref{Fig:particle_hierarchy}, the hierarchical particle structure consists of three layers: the third layer is the original particle set for belief update, which maintains sufficient particles as an adequate representation of belief.
Simplified through the fine grid, particles of the second layer are used to compute the uncertainty measure with improved efficiency.
The first layer consists of high-level particles that cluster through coarse grid subdivisions, which guides the robot to execute more reasonable trajectories capitalizing on global estimation information.
The hierarchical particle structure serves dual purposes.
First, in scenarios where belief uncertainty is significant, characterized by dispersed particles, the hierarchical structure can extract critical particles to direct the robot towards a planning scheme that considers both the global path efficiency and the local uncertainty reduction.
Second, while ensuring the efficacy of the planning process, the hierarchical structure can flexibly adjust the number of particles to enable efficient MI computation and belief update in the planning module. 

\begin{algorithm}[!t]
\caption{Particle Hierarchy}
\fontsize{10pt}{10pt}\selectfont
\label{alg:H_Particle}
\textbf{procedure} \textsc{ParticleHierarchy}($\boldsymbol{B}$)\\
$\boldsymbol{B}_{h} \leftarrow \textsc{CoarseGrid}(\boldsymbol{B},l_c)$\\
$\boldsymbol{B}_{goal} \leftarrow \textsc{FindGoal}(\boldsymbol{B}_{h})$\\
$\boldsymbol{B}_{c} \leftarrow \textsc{LocalParticle}(\boldsymbol{B}_{goal})$\\
$\boldsymbol{B}_{s} \leftarrow \textsc{FineGrid}(\boldsymbol{B}_c,l_f)$\\
$\widetilde{\boldsymbol{B}} = \{\boldsymbol{B}_c,\boldsymbol{B}_s\}$\\
\Return $\widetilde{\boldsymbol{B}}$
\end{algorithm}

The main steps of particle hierarchy are shown in \Cref{alg:H_Particle}.
The algorithm first decomposes the state space with a coarse grid that contains cells with size $l_c$, and extracts high-level particles $\boldsymbol{B}_{h}$ by calculating the weighted average of particles within each cell (line 2).
Regarding high-level particles as waypoints, the algorithm finds the shortest path that starts from the current robot position and traverses these waypoints considering present obstacle information, and designates the subsequent waypoint $\boldsymbol{B}_{goal}$ along this route as the goal point (line 3).
Next, the original particles corresponding to the goal point form the \textit{critical particles} $\boldsymbol{B}_{c}$ (line 4), and subsequently the simplified version $\boldsymbol{B}_{s}$ of critical particles is obtained by partitioning the space regularly with grid size $l_f$ and replacing particles in the same grid with their weighted average (line 5).
Finally, the process outputs the critical particles and simplified particles as the belief input for subsequent planning procedures.
Such a hierarchical approach can reduce the computational resources allocated to particle update in the planning module while guaranteeing reasonable decision-making capabilities.




\subsection{Reusable Belief Tree Search \label{subsec:RCTS}}

This subsection proposes the RBTS, a novel tree search method that explores the belief space for planning under uncertainty.
Compared to the standard MCTS method, RBTS has two main improvements. 
First, unlike the vanilla tree search method that ignores the influence of sensing uncertainty and only searches for the open-loop strategy, our policy tree expands action nodes and belief nodes by simulating the action-observation sequence to account for the sensing uncertainty. 
Second, we develop a new recycling process to substitute the simulation step in the standard method, which utilizes the existing rollout evaluations to reduce computational overhead.

\subsubsection{Algorithm Overview}

\begin{figure}[!t] 
\centering
\includegraphics[width=0.49\textwidth]{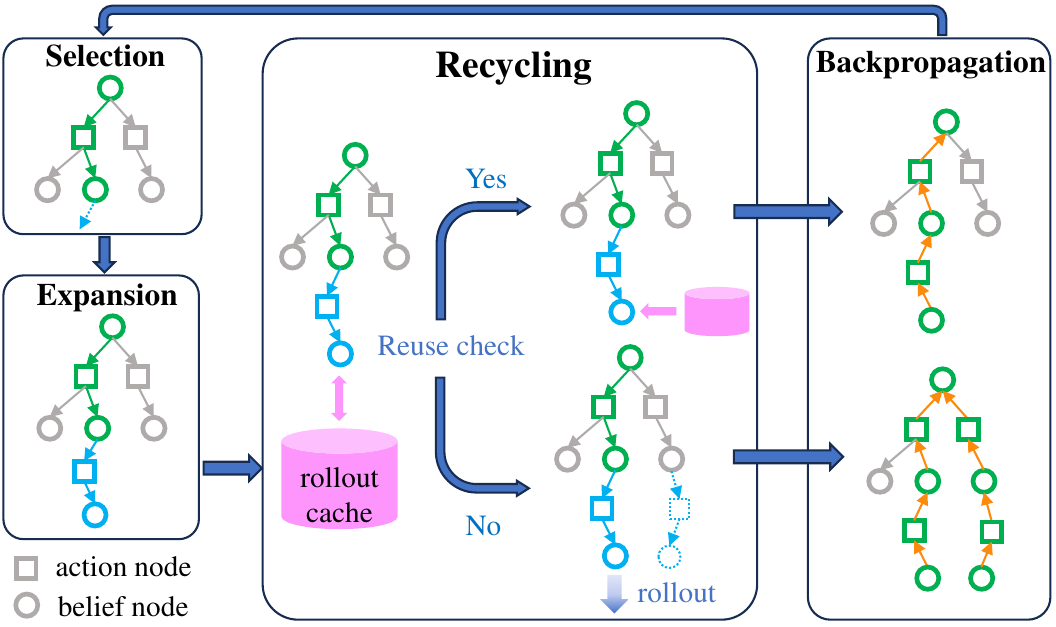} 
\caption{\textbf{Illustration of RBTS.}
A novel recycling step is introduced to leverage past information stored in the rollout cache, thereby simultaneously expanding multiple nodes to enhance planning efficiency.} 
\label{Fig:MCTS_recycle} 
\end{figure}

The RBTS comprises four steps: selection, expansion, recycling, and backpropagation, which is illustrated in \Cref{Fig:MCTS_recycle}.
By iterating through these steps, we construct a policy tree consisting of action nodes and belief nodes for planning under sensing uncertainty (see \Cref{Alg:policy_tree}).
The information stored in the node $n$ consists of action set $\mathcal{A}_n$, action-observation history $\mathcal{H}_n$, belief state $\boldsymbol{B}_n$, children set $C_n$, and visit counts $W_n$.
Additionally, the action node contains the accumulated value $Q_n$ for action selection. 

Given the current map $\mathcal{M}$ and root node $n_r$ created with current belief, the policy tree $\mathcal{T}$ is first initialized (Line 2).
Subsequently, each iteration starts by selecting a leaf node within the current policy tree that possesses unexpanded child nodes, following the upper confidence bound (UCB) criterion~\cite{kocsis2006bandit} (Line 4).
After a leaf node is chosen, by sampling the action-observation pair ($\boldsymbol{a},\boldsymbol{o}$),
the expansion step updates belief with the particle filter and generates a new belief node $n_{new}$ for tree growth (Line 5-6).
We pre-define the robot action space that consists of motion primitives generated by the robot kinematics, and the action is selected from the action space to allow for smooth trajectory execution.

Next, a recycling step is called to estimate the long-term reward of the newly expanded node (Line 7), which reuses the existing rollout evaluation for efficiency enhancement.
The procedure will be detailed in \Cref{subsec:recycle}.
Last, the information in visited nodes is updated with the rollout evaluation in the backpropagation step similar to~\cite{best2019dec} (Lines 8-9).

\subsubsection{Recycling for Rollout Reuse}
\label{subsec:recycle}

\begin{algorithm}[!t]
\caption{Reusable Belief Tree Search}
\fontsize{10pt}{10pt}\selectfont
\label{Alg:policy_tree}
\textbf{procedure} \textsc{BuildTree}($\mathcal{M},n_r$)\\
$\mathcal{T} \leftarrow n_r$, $\mathcal{S} \leftarrow \varnothing$\\
\While{number of tree nodes $\leq N_{max}$ }{
$n \leftarrow$ \textsc{Selection}($\mathcal{T}$)\\
$(\boldsymbol{a},\boldsymbol{o}) \leftarrow$ \textsc{Sampling}($n$)\\
$n_{\rm new} \leftarrow$ \textsc{Expansion}($\mathcal{M},n,\boldsymbol{a},\boldsymbol{o}$)\\
$\mathcal{T},\mathcal{S},\mathcal{S}_{\rm new} \leftarrow$ \textsc{Recycling}($\mathcal{T},\mathcal{S},n_{\rm new}$)\\
\ForAll{$n$ in $\mathcal{S}_{\rm new}$}{
$\mathcal{T} \leftarrow \textsc{Backpropagation}(\mathcal{T},Q(n))$\\}}
\Return\\

\end{algorithm}

\begin{algorithm}[!t]
\caption{Recycling Procedure}
\fontsize{10pt}{10pt}\selectfont
\label{Alg:recycle}
\textbf{procedure} \textsc{Recycling}($\mathcal{T},\mathcal{S},n$)\\
$(\mathcal{C}_{min}, d_{min}) \leftarrow \textsc{ClosestCluster}(\mathcal{S},n)$\\
\If{$d_{min} \leq d_{thr}$}{
$Q_n \leftarrow \mathcal{C}_{min}.r$ \\
$\mathcal{S}_{\rm new} \leftarrow \{n\}$\\
\Return $\mathcal{T},\mathcal{S},\mathcal{S}_{\rm new}$}
$Q_n \leftarrow$ \textsc{Simulation}($\mathcal{T},n$)\\
$\mathcal{S} \leftarrow \mathcal{S} \cup (n,Q_n)$\\
$\mathcal{S}_{\rm new} \leftarrow \{n\}$\\
$\boldsymbol{o} \leftarrow$ last observation from $\mathcal{H}_n$\\
\For{$n_b$ in $\mathcal{T}$}{
\For{$\boldsymbol{a}$ in $\mathcal{A}_{n_b}$}{
$n_{\rm new} \leftarrow$ \textsc{Expansion}($\mathcal{M},n_b,\boldsymbol{a},\boldsymbol{o}$)\\
\If{$d(n,n_{new}) \leq d_{thr}$}{
$Q_{n_{\rm new}} \leftarrow Q_n$\\
$\mathcal{S}_{\rm new} = \mathcal{S}_{\rm new} \cup n_{\rm new}$\\}}
}
\Return $\mathcal{T},\mathcal{S},\mathcal{S}_{\rm new}$\\
\end{algorithm}

In the traditional simulation step, a rollout procedure is called to estimate the accumulated reward of the newly expanded node.
However, due to the computationally intensive nature of information-theoretic reward calculation, which is iteratively executed throughout the rollout process, the computational cost associated with the rollout is substantial.
We introduce a recycling procedure to reuse rollout evaluation and enhance the tree search efficiency.

Concretely, we define a structure called \textit{rollout cluster} $\mathcal{C}=\{n_b,r\}$ including the belief node $n_b$ that rollout applies to, and the associated rollout reward $r$.
During the construction of the policy tree, we maintain and update a \textit{rollout cache} $\mathcal{S}$, which stores existing rollout clusters to provide potential reuses for subsequent rollouts.
\Cref{Alg:recycle} gives the process of recycling procedure. 
Given the newly expanded node $n$ and rollout cache $\mathcal{S}$, the algorithm first finds the closest cluster $\mathcal{C}_{min}$ and corresponding distance $d_{min}$ in the cache based on the distance function $d(n_1,n_2)$ between two nodes $n_1$ and $n_2$,
\begin{equation}
\begin{split}
d(n_1,n_2)=\begin{cases}
|\boldsymbol{x}_1-\boldsymbol{x}_2| & |\boldsymbol{o}_1-\boldsymbol{o}_2|\leq \boldsymbol{o}_{thr}\\
\infty & |\boldsymbol{o}_1-\boldsymbol{o}_2|> \boldsymbol{o}_{thr}
\end{cases},
\end{split}
\end{equation}
where $\boldsymbol{x_1}$ and $\boldsymbol{x}_2$ are the robot state from $\boldsymbol{B}_{n_1}$ and $\boldsymbol{B}_{n_2}$, $\boldsymbol{o_1}$ and $\boldsymbol{o}_2$ are the last observations from $\mathcal{H}_{n_1}$ and $\mathcal{H}_{n_2}$, and $\boldsymbol{o}_{thr}$ is the specified observation similarity threshold.
If the closest distance $d_{min}$ is less than the specified threshold $d_{thr}$, rollout reuse is carried out and the recycling procedure terminates (Lines 3-6).
Otherwise, the simulation step is conducted, where the rollout is applied to the newly expanded node, and the information is stored in the rollout cache (Lines 7-8).
After the rollout is completed, we retrieve all belief nodes in the current policy tree to generate child node $n_{new}$ similar to $n$, satisfying $d(n_{new},n)<d_{thr}$, and reuse the rollout reward (Lines 11-16).
Moreover, all newly expanded nodes are stored in $\mathcal{S}_{\rm new}$ for subsequent backpropagation. 
Note that the standard tree search method typically expands one node in the expansion procedure, but our method can expand multiple nodes in one iteration, improving the efficiency of the tree growth.

\begin{figure*}[t]
\centering
\includegraphics[width=0.98\linewidth]{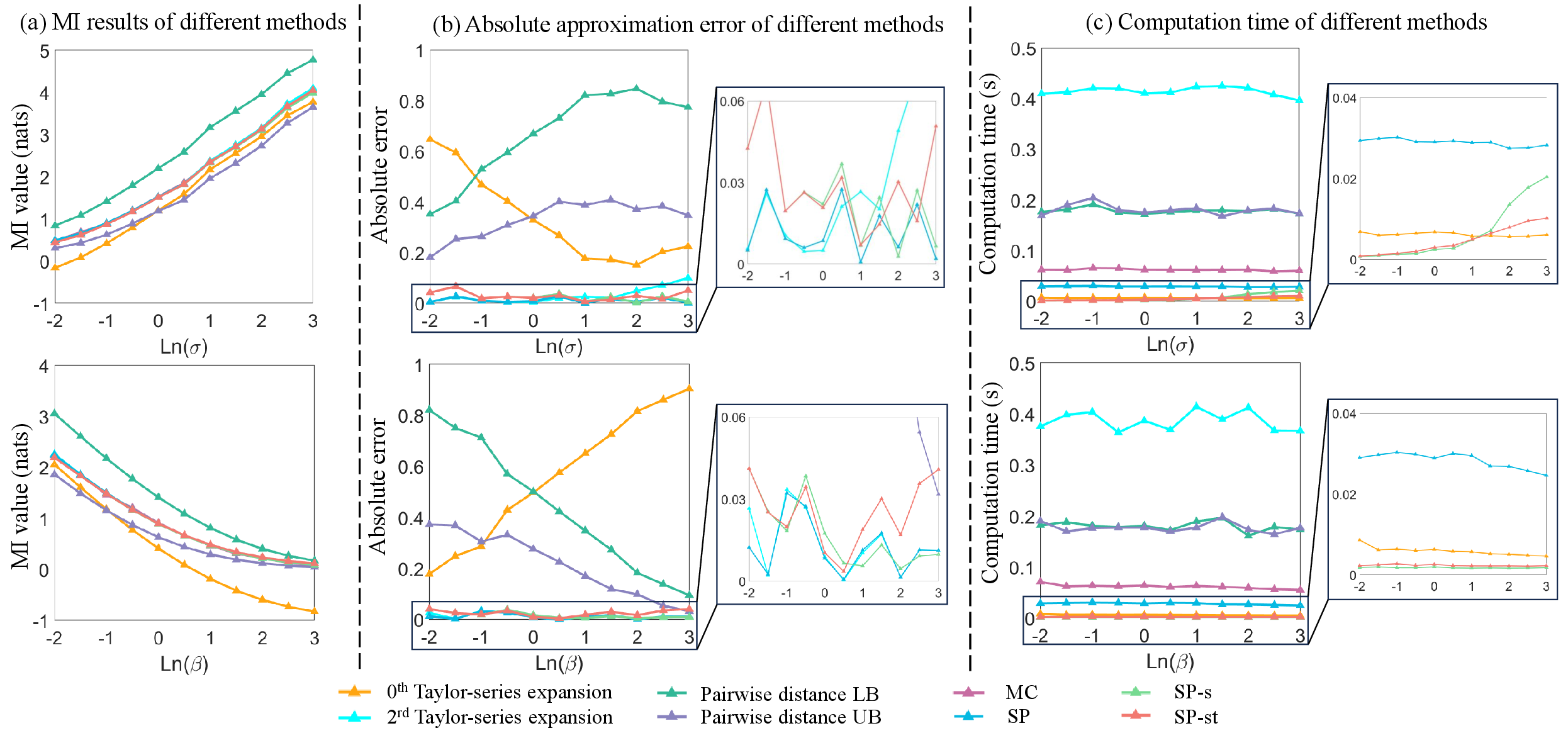}
\caption{\textbf{Comparison of the MI approximation considering particle dispersion (top row) and sensor noise (bottom row).}
MI result (left column), absolute approximation error (middle column), and computation time (right column) are displayed.
The absolute error and computation time are zoomed in for clear visualization.
}
\label{Fig:MI_comparison_result}
\end{figure*}

\section{Simulations and Discussion}

We conduct extensive simulations to demonstrate the effectiveness of the proposed method in
MATLAB using a desktop with Intel Core i7 CPU@2.10GHz and 16GB RAM. 
First, the SP-based approximation method is compared with other baselines to illustrate the approximation accuracy and computational efficiency. 
Subsequently, we compare ReSPIRe with several representative benchmarks commonly utilized in SAT tasks in unknown cluttered environments. 
Finally, an ablation study is conducted to further illustrate the benefits of the hierarchical structure and rollout reuse.

\subsection{Comparison of MI Approximation Methods
\label{subsec:MI_comparison}}

To evaluate our SP-based approximation method, we compare the performance of a series of entropy approximation methods in estimating the GMM entropy.
The proposed methods include the vanilla SP-based approximation method, SP-based approximation method with particle simplification~\cite{charrow2014approximate}, and SP-based approximation method with both particle simplification and truncation, referred to as SP, SP-s, SP-st, respectively. 
The benchmarks include three typical entropy approximation approaches for mixture distributions, including Taylor-series expansion~\cite{huber2008entropy}, Pairwise distance~\cite{kolchinsky2017estimating}, and Monte Carlo integration, referred to as MC.

We generate $500$ particles as potential target states and compute the entropy of the observation derived from the particles.
We conduct two types of simulations, where the first investigates the impact of varying degrees of particle dispersion on the approximation results, and the second evaluates the performance of these methods under different sensor noise.
We choose the MI value, absolute error, and computation time as the metrics.
Regarding the MC result as the ground truth, the absolute error is defined as the absolute deviation from the MC result.
We consider the following range-bearing sensor as the observation model, 
\begin{equation}
\begin{split}
\mathbf{h}(\boldsymbol{x}_{k}^{r},\boldsymbol{x}_{k}^{t}) = \begin{bmatrix}
\sqrt{({x}_{k}^{r} - {x}_{k}^{t})^2 + ({y}_{k}^{r} - {y}_{k}^{t})^2} \\
\arctan({y}_{k}^{t} - {y}_{k}^{r}, {x}_{k}^{t} - {x}_{k}^{r}) - \theta_k^r
\end{bmatrix}.
\end{split}
\end{equation}
where ${x}_{k}^{r}$, ${y}_{k}^{r}$ and $\theta_k^r$ denote x-y position and angle of the robot, respectively, and ${x}_{k}^{t}$, ${y}_{k}^{t}$ denote the target's position.

\subsubsection{Particle dispersion}

In the first simulation, we set the covariance of sensor noise as $diag(0.1,0.01)$.
The robot state is set as ${[0, 0, 0]}^T$, and particles are generated by sampling from a Gaussian distribution $\mathcal{N}({[10,0]}^T,\alpha I_2)$ with varying covariance, where $I_d$ represents the $d\times d$ identity matrix.
As depicted in \cref{Fig:MI_comparison_result}(a) and \cref{Fig:MI_comparison_result}(b), our SP-based approximation method and its variants yield more accurate entropy estimation results compared to other baselines in different dispersion levels, exhibiting lower absolute errors.
Albeit the $2^{nd}$ Taylor-series expansion also achieves high estimation accuracy, it requires the longest computation time.
Moreover, we can notice from \cref{Fig:MI_comparison_result}(c) that when particles cluster closely, combined with the particle simplification technique, SP-s significantly enhances computational efficiency.
As the particles disperse, the computation time of SP-s increases due to the diminished impact of simplification.
When further integrated with the truncation method, SP-st can reduce computation time by $50\%$, as indicated in \cref{Fig:MI_comparison_result}(d).

\subsubsection{Sensor noise}

In the second simulation, we set the robot state as ${[0, 0, 0]}^T$, and generate particles by sampling from a fixed Gaussian distribution $\mathcal{N}({[10,0]}^T,I_2)$.
We set the covariance of sensor noise as $\beta \boldsymbol{\Phi}$, where $\boldsymbol{\Phi} = diag(0.1,0.01)$, and adjust the sensor noise by modifying the parameter $\beta$.
Consistent with the first simulation, the proposed series of SP-based methods obtain precise estimation as shown in \Cref{Fig:MI_comparison_result}(d) and \Cref{Fig:MI_comparison_result}(e).
The particle simplification and truncation technique again provide significant computational efficiency improvement in \Cref{Fig:MI_comparison_result}(f).
Note the different trend of computation time in \Cref{Fig:MI_comparison_result}(f) compared to \Cref{Fig:MI_comparison_result}(c).
This inconsistency arises from the unchanged particle distribution in the second simulation, which results in the same efficiency improvement achieved by the simplification method under different sensor noise.
In summary, these results prove the advantages of the proposed approaches in approximation accuracy and computational efficiency.

\subsection{Comparison of Planning Methods}

To validate the superiority of the proposed planning method, we compare the proposed method against several representative baselines to evaluate their performance in SAT tasks under unknown cluttered environments and large prior uncertainty.
We design two $50 \times 50$ $m^2$ planar spaces comprising multiple obstacles as the simulation environment, depicted in the leftmost column of \Cref{Fig:qualitative_result}.
To evaluate the generalization capability, we create $10$ scenarios where the initial poses of the robot and the target, and target trajectories are randomly generated. 
The initial distance between the robot and the target is sufficiently far to guarantee the complexity of the task.
Each scenario lasts for 200 simulation steps.

\begin{table}[t]
    \centering
    \caption{Simulation parameters in comparison of planning methods}
    \small
    \begin{tabular}{@{}lc@{}}
        \toprule
        \textbf{Parameters} & \textbf{Value} \\ \midrule
        Particle number & 500 \\
        Planning horizon (search) & 10 \\
        Planning horizon (tracking) & 5 \\
        Number of tree nodes ($N_{max}$) & 100 \\
        Grid size ($l_c$) & 10 \\
        Observation similarity threshold ($\boldsymbol{o}_{th}$)& 0.1 \\ 
        Robot linear velocity ($v_{k}^{r}$) & $[0,3]$ m/s \\
        Robot angular velocity ($w_{k}^{r}$) & $[-\pi/3,\pi/3]$ rad/s \\
        Sampling interval ($\Delta t$) & 0.5 \\
        \bottomrule
    \end{tabular}
    \label{tab:simulation_parameters}
\end{table}

We consider three types of baselines: the first type that only replaces ReSPIRe with other planning methods and keeps the other parts of the framework in \cref{Fig:overview} unchanged, and the second type that replaces the entire framework.
Specific methods are explained as follows:
\begin{itemize}
\item The first type of baselines consists of representative information-gathering planners, including \textbf{NBV}, \textbf{IIG-tree}~\cite{ghaffari2019sampling}, a sampling-based informative planner, and PFT~\cite{sunberg2018online}, an online POMDP solver adapted to belief-dependent reward, \textbf{GVF}~\cite{kong2023collision}, and \textbf{CBF}~\cite{li2024collision}, which respectively utilize guidance vector field and control barrier function for collision-free target seeking.
In PFT, the particle number significantly influences the balance between algorithm effectiveness and efficiency, so we consider PFT with $50$ and $500$ particles, named \textbf{PFT-}$\mathbf{50}$ and \textbf{PFT-}$\mathbf{500}$, to investigate the performance of PFT with different particle number.

\begin{figure}[t]
\centering
\includegraphics[width=\linewidth]{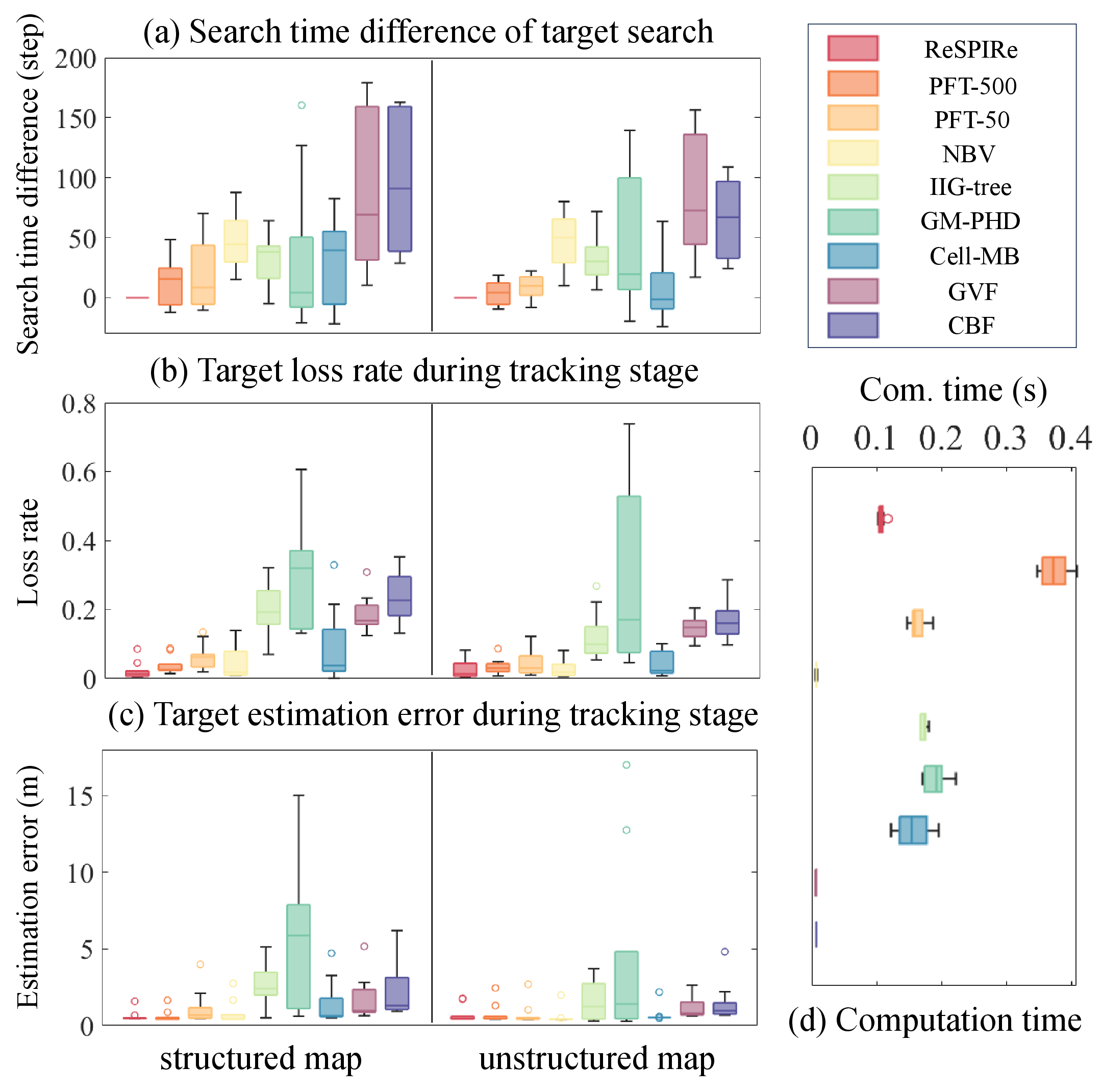}
\caption{\textbf{Quantitative comparison in SAT task.}
The left column and right column in (a)-(c) represent the results in structured map and unstructured map, respectively. 
(d) illustrates the computation time of the planning module for each method throughout the entire simulation including both structured and unstructured maps.
}
\label{Fig:simulation_quantitative_result}
\end{figure}

\begin{figure*}[t]
\centering
\includegraphics[width=\linewidth]{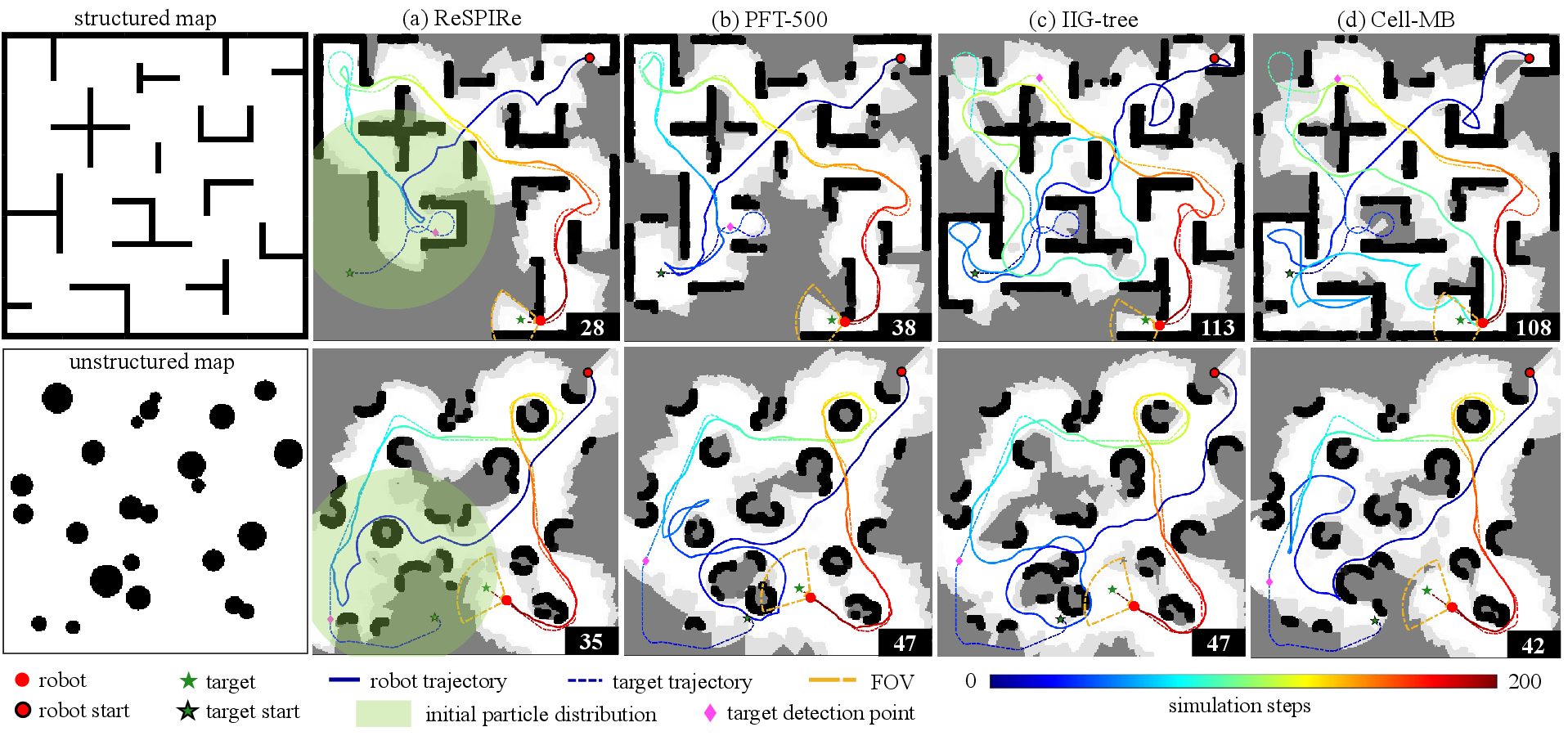}
\caption{\textbf{Trajectory comparison in structured map (top row) and unstructured map (bottom row).}
The colormap illustrates different simulation steps of the trajectories.
The pink diamond marks the target position where the robot finds the target.
The number in the bottom right corner of each subfigure indicates the simulation steps spent to find the target for each approach.
}
\label{Fig:qualitative_result}
\end{figure*}

\item The second type of baselines consists of \textbf{GM-PHD} filter~\cite{sung2021gm} and \textbf{Cell-MB} sensor control~\cite{legrand2022cell}.
These methods have developed their own estimation techniques.
Therefore, comparing with these baselines allows evaluation of the proposed complete framework for SAT.
\end{itemize}
The robot motion models use the following unicycle model,
\begin{equation}
\mathbf{f}^r(\boldsymbol{x}_{k}^r,\boldsymbol{u}_{k}^r)=\boldsymbol{x}_{k}^r+[v_{k}^r\cos\theta_{k}^r, v_{k}^r\sin\theta_{k}^r, w_{k}^r]^T \cdot \Delta t,
\end{equation}
where $v_{k}^r$ and $w_{k}^r$ denote the robot's velocity and angular velocity, respectively, and $\Delta t$ is the sampling interval. 
The simulation parameters are listed in \Cref{tab:simulation_parameters}.
Robot velocity limits are determined by environmental scale and hardware constraints, while other parameters are empirically chosen to balance computational efficiency and planning performance.
The maximum sensing range is set as $6~m$, and the sensing angle for the camera and lidar is $90^\circ$.
Four metrics are evaluated for quantitative comparisons: the search time difference $t_s$, target loss rate $r_{los}$, estimation error $\varepsilon_{est}$, and computation time.
To compare the search efficiency in each scenario, we define $t_s$ as the difference in simulation steps to find the target compared with ReSPIRe.
The target loss rate is defined as $r_{los} = \frac{T_{los}}{T_{tra}}$, where $T_{los}$ is the target loss time during the tracking stage, and $T_{tra}$ is the total tracking time. 
The estimation error is defined as $\varepsilon_{est}=\frac{1}{T_{tra}} \sum_{k=1}^{T_{tra}} ||\boldsymbol{x}_{k}^{t}-\hat{\boldsymbol{x}}_{k}^{t}||$, where $\hat{\boldsymbol{x}}_{k}^{t}$ is the estimation of the target positions. 
For each scenario, we repeat $10$ trials and average the results, which are shown in \Cref{Fig:simulation_quantitative_result}.

\begin{table*}[htbp]
  \centering
  \resizebox{0.98\linewidth}{!}{
  \begin{tabular}{|c|c|c|c|c|c|c|c|c|c|c|c|}
    \toprule
    \multirow{2}{*}{Methods}& \multirow{2}{*}{Com. time (s)} & \multicolumn{10}{|c|}{Search time (step)}\\
    \cmidrule(l{1pt}r{1pt}){3-12}
    && scenario 1 & scenario 2 & scenario 3 & scenario 4 & scenario 5 & scenario 6 & scenario 7 & scenario 8 & scenario 9 & scenario 10\\
    \midrule
    Van & $0.481\pm0.031$ & $42.7\pm17.6$ & $75.6\pm40.5$ & $56.4\pm17.2$ & $65.9\pm31.0$ & $\textbf{36.7}\pm\textbf{25.9}$ & $50.4\pm26.5$ & $32.1\pm21.1$ & $52.3\pm27.6$ & $35.6\pm4.9$ & $46.1\pm8.5$\\
    \midrule
    Van+R & $0.210\pm0.015$ & $54.3\pm15.4$ & $56.8\pm46.8$ &$63.8\pm15.0$ & $79.4\pm38.4$ & $45.9\pm25.0$ & $53.2\pm19.3$ & $42.5\pm29.4$ & $49.4\pm25.3$ & $39.5\pm5.8$ & $48.9\pm20.8$\\
    \midrule
    Van+H & $0.221\pm0.018$ & $\textbf{35.3}\pm\textbf{3.7}$ & $34.5\pm18.4$ & $\textbf{37.5}\pm\textbf{19.1}$ & $34.9\pm25.7$ & $45.6\pm18.9$ & $40.2\pm19.3$ & $29.6\pm9.6$ & $\textbf{32.0}\pm\textbf{9.4}$ & $38.7\pm16.5$ & $45.8\pm14.7$\\
    \midrule
    ReSPIRe & $\textbf{0.102}\pm\textbf{0.009}$ & $36.1\pm10.5$ & $\textbf{31.9}\pm\textbf{3.7}$ & $38.8\pm11.1$ & $\textbf{29.8}\pm\textbf{9.2}$ & $42.1\pm13.5$ & $\textbf{36.0}\pm\textbf{11.9}$ & $\textbf{29.2}\pm\textbf{11.5}$ & $34.9\pm8.4$ & $\textbf{34.7}\pm\textbf{9.3}$ & $\textbf{41.1}\pm\textbf{13.1}$\\
    \bottomrule
  \end{tabular}}
  \caption{Ablation study of the hierarchical particle structure and rollout reuse.
  }
  \label{tab:ablation_study}
\end{table*}

As depicted in \Cref{Fig:simulation_quantitative_result}, ReSPIRe exhibits distinct advantages across all four metrics over baselines in two different maps.
In the search time comparison presented in \Cref{Fig:simulation_quantitative_result}(a), the majority of the boxplots for all baselines are above zero, which indicates that ReSPIRe outperforms the baselines in terms of search efficiency across a multitude of scenarios.
\Cref{Fig:simulation_quantitative_result}(b) reveals that ReSPIRe has the lowest target loss rate, which corresponds to the lowest estimation error displayed in \Cref{Fig:simulation_quantitative_result}(c), demonstrating that ReSPIRe also yields superior and stable performance in tracking tasks.

In comparison to promising results of ReSPIRe, baseline methods show poor performance in SAT tasks due to various limitations.
Adhering to simple planning strategies, where NBV and Cell-MB employ the greedy policy, and GM-PHD directly moves to the closest mean of the Gaussian component, these methods exhibit diminished planning performance in the intricate SAT tasks. 
Due to the random nature of samples, IIG-tree generates sinuous trajectories and usually loses sight of the target, leading to high estimation error.
While GVF and CBF are computationally efficient due to their analytical control strategies, constrained by the assumption of known circular obstacles, they often fail in environments with unknown or concave obstacles. 
Their relatively simple strategies also struggle to achieve a reasonable trade-off between stable target tracking and obstacle avoidance.
Conversely, ReSPIRe utilizes the hierarchical particle structure to extract critical particles from dispersed distribution, guiding effective global planning route under considerable uncertainty.
Moreover, ReSPIRe leverages the online tree search to explore non-myopic trajectories, facilitating a more farsighted approach to long-term planning.

Furthermore, as indicated in \Cref{Fig:simulation_quantitative_result}(d), ReSPIRe can achieve a computation frequency of 9.7 Hz, ensuring the real-time operational capability.
In contrast, the complex objective function evaluations impose additional computational costs on Cell-MB and GM-PHD, while IIG-tree necessitates numerous samples to ensure planning quality, thereby incurring further computational overhead.
PFT-500 achieves the best performance among the baselines, but the extensive particles significantly inflate the computational demands for MI calculation and belief transition, rendering real-time planning impractical. 
Conversely, PFT-50 utilizes fewer particles and uses shorter computation time, but at the expense of a less expressive belief representation, diminishing its estimation accuracy and subsequent planning performance in SAT tasks. 
In contrast to the fixed number of particles in these PFT methods, ReSPIRe enables flexible adjustment of the particle counts with hierarchical particle structure.
Coupled with the reusable tree search, ReSPIRe enhances algorithm efficiency without compromising the quality of the planning results.
In contrast to the fixed number of particles in these PFT methods, ReSPIRe enables flexible adjustment of the particle counts with hierarchical particle structure.
Coupled with the reusable tree search, ReSPIRe enhances algorithm efficiency without compromising the quality of the planning results.

\Cref{Fig:qualitative_result} shows the trajectories generated by four methods that perform reasonably well in the tasks.
We can see that the baselines meander in the dispersed particle distribution and revisit some areas, indicating redundant behavior to find the missing target in unknown cluttered environments.
In contrast, benefitting from the guidance of the critical particles obtained from particle hierarchy, ReSPIRe demonstrates higher search efficiency in complex SAT tasks with considerably uncertain prior target information.
Furthermore, with continuous extension of motion primitives in the policy tree, ReSPIRe can track the target with more rational and smoother trajectories.

\subsection{Ablation Study}

To further investigate the effectiveness of the hierarchical particle structure and rollout reuse, we conduct ablation study to compare ReSPIRe with three variants.
The first variant \textit{Van} is the vanilla MCTS-based approach without particle hierarchy and rollout reuse.
The other two variants \textit{Van+R} and \textit{Van+H} are vanilla methods combined with rollout reuse and particle hierarchy, respectively.
Considering the randomness in dynamic target search, where the non-adversary target may inadvertently move to the target, the ablation study focused on static targets. 
To maintain the search difficulty of the task, we designate several challenging-to-detect corners within the structured map as the possible target positions.
We design $10$ scenarios, where the target is initialized at these challenging positions randomly, and the robot's initial position is sampled from the free workspace.
We repeat $10$ trials for each scenario and record the search time and computation time.

\begin{figure}[t]
\centering
\includegraphics[width=0.98\linewidth]{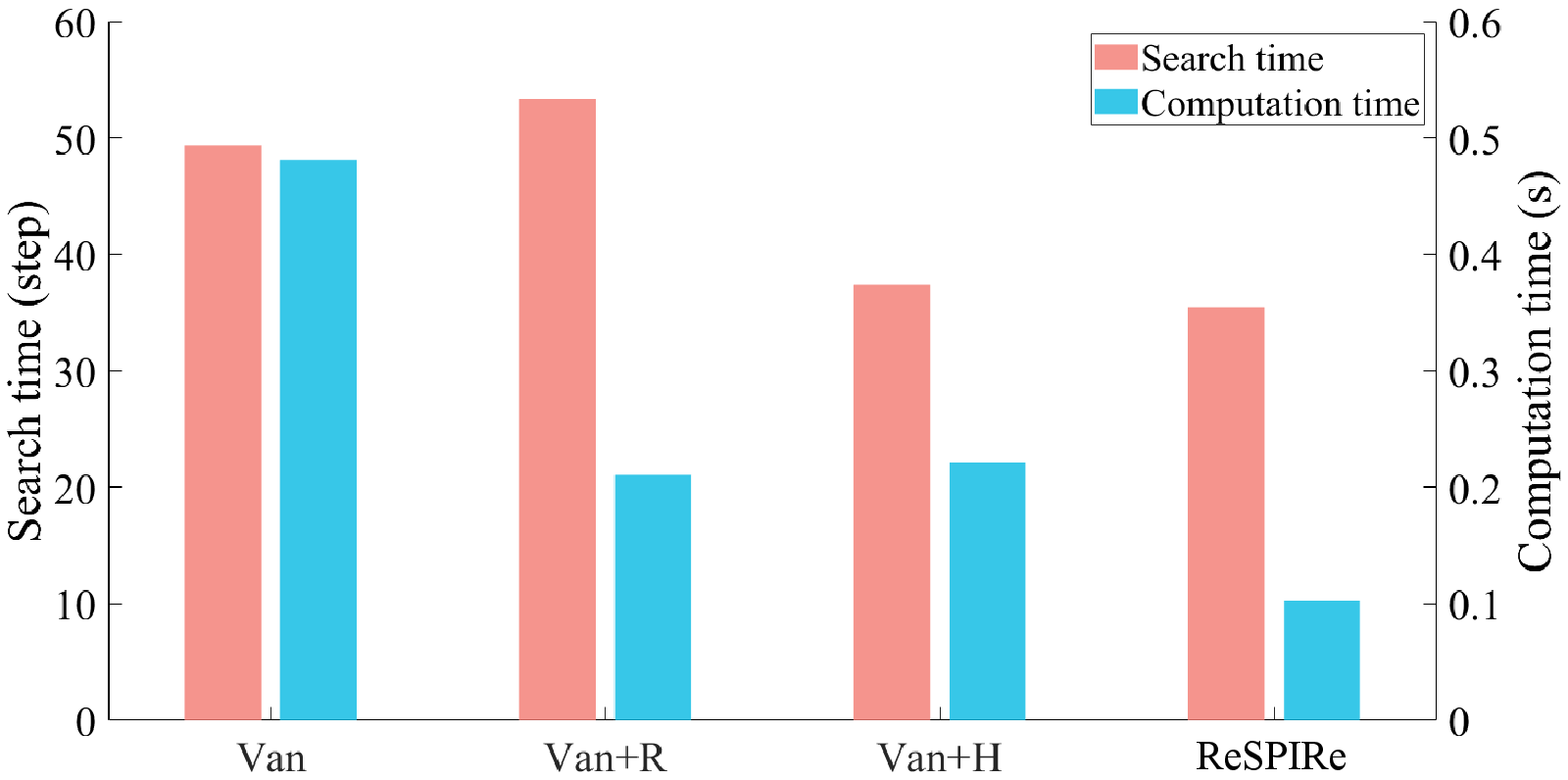}
\caption{
\textbf{Average search time and computation time in ablation study.}
}
\label{Fig:ablation_study_comparison}
\end{figure}

\begin{figure*}[t]
\centering
\includegraphics[width=\linewidth]{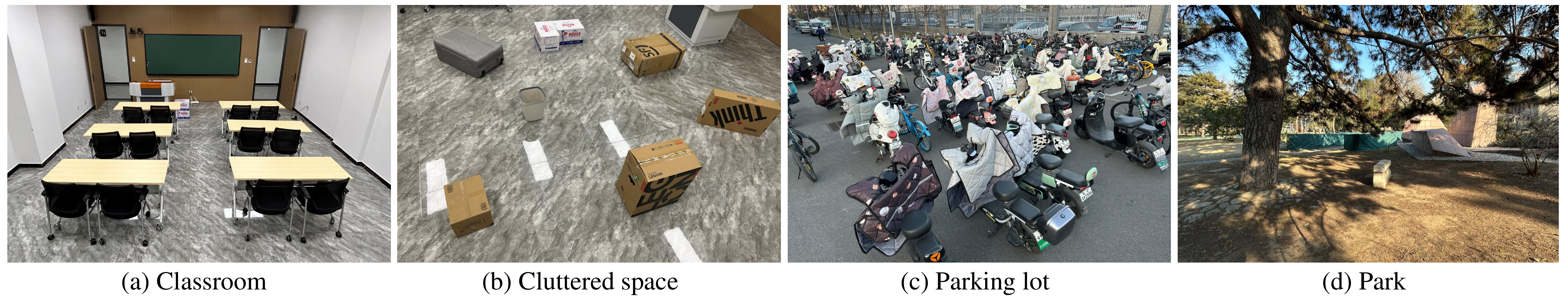}
\caption{\textbf{Real-world experimental scenarios.}
(a) is a classroom with crowded desks and chairs and (b) is a cluttered space with pervasive obstacles.
(c) and (d) are outdoor scenarios, where (c) is a crowded bicycle parking lot and (d) is a park with irregular terrains.
}
\label{Fig:exp_environment}
\end{figure*}

As depicted in \Cref{tab:ablation_study}, introducing particle hierarchy can obtain a shorter search time in 9 scenarios out of 10, while enhancing the computational efficiency. 
The only exception observed in scenario 5 can be attributed to the stochastic nature of the scenario setup, which inadvertently simplified the search task, leading to insignificant performance disparities among all methods.
Furthermore, note that though van+H and ReSPIRe have their own strengths in terms of search time, ReSPIRe is notably faster in computation time.
\Cref{Fig:ablation_study_comparison} shows the average search time and computation time of all trials in the ablation study, and we can conclude that the particle hierarchy improves search efficiency by $20\%$-$30\%$ and computational efficiency by $60\%$, and the rollout reuse enhances the computational efficiency by $60\%$ with little compromise on the target search speed.
Moreover, the standard deviation of other baselines in the ablation study is generally larger than that of ReSPIRe, demonstrating the stability of our approach across various scenarios.
These results demonstrate the effectiveness of the particle hierarchy and rollout reuse.

\section{Real-World Experiments}

We also conducted indoor and outdoor experiments to verify the performance of ReSPIRe in real-world scenarios.
The experimental environments include two indoor and two outdoor scenarios, as shown in \Cref{Fig:exp_environment}, encompassing various realistic and complex environments with cluttered obstacles and irregular terrain.

\begin{figure}[t]
\centering
\includegraphics[width=\linewidth]{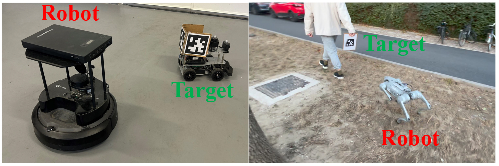}
\caption{\textbf{The Robot and target in real-world experiments.}
We use a Turtlebot4 as the robot and a Wheeltec R550 Ackermann ground robot as the target in indoor experiments, and we utilize the Unitree Go2 as the robot to search for and track a person in outdoor experiments.
All targets carry an Apriltag for detection.
}
\label{Fig:exp_robot}
\end{figure}

To demonstrate the generalizability of the proposed method across different robotic platforms, we adopted a Turtlebot4 wheeled robot for indoor experiments and a Unitree Go2 quadruped robot for outdoor experiments, as depicted in \Cref{Fig:exp_robot}.
To simplify the sensing module, the target carries an Apriltag for target detection.
We implemented ReSPIRe using C++ code for efficient execution in realistic robotic platforms and modified the related parameters to achieve a better balance between computational efficiency and algorithm effectiveness in experiments.
We also implemented PFT-50~\cite{sunberg2018online} by C++ code, which demonstrated promising performance among baselines in the simulations, for real-world experimental comparison.
We conducted three repeated experiments in each scenario for reproducibility validation, with each experiment lasting at least 150 seconds.

\subsection{Indoor Experiments}

\begin{figure*}[t]
\centering
\includegraphics[width=\linewidth]{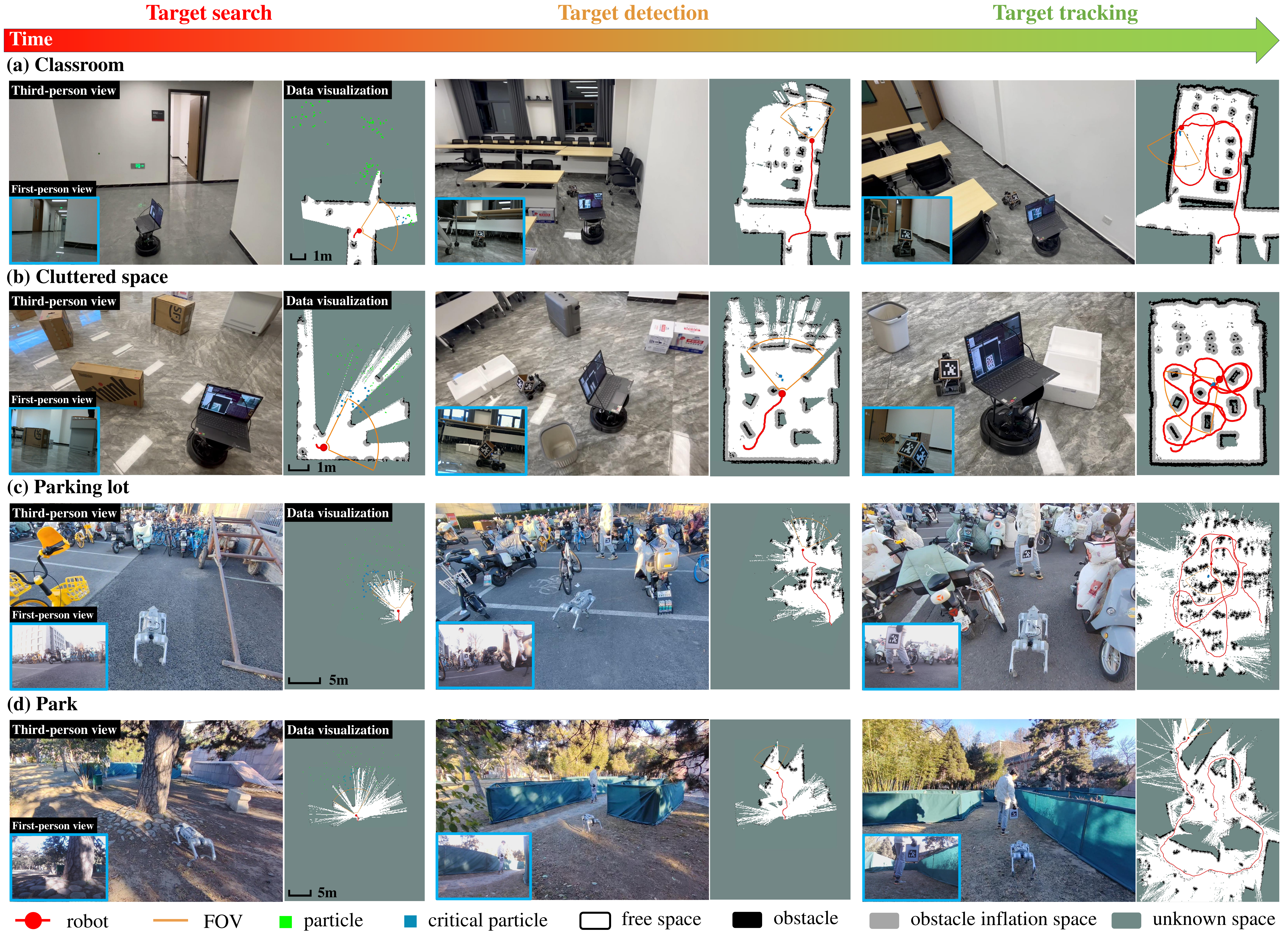}
\caption{\textbf{SAT process in real-world experiments.}
Each row illustrates the SAT process for one scenario, with three horizontally arranged figures displaying three stages: target search, detecting the target, and target tracking. 
In each subfigure, the left picture displays the third-person and first-person perspectives in the realistic scenario, while the right picture visualizes the robot's trajectory, map, and particles.
(a) A classroom and outside corridor. 
(b) A cluttered space with various obstacles.
(c) A crowded bicycle parking lot.
(d) A large park with irregular terrains and unstructured obstacles.
We refer the reader to the attached video for the complete SAT process in diverse scenarios.
}
\label{Fig:exp_process}
\end{figure*}

In indoor experiments, we used the Turtlebot4 as the robot and a Wheeltec R550 Ackermann robot as the target to complete the SAT tasks.
The Turtlebot4 carries an RPLidar-A1 for 2d mapping, a RealSense depth camera D435i for visual detection, and a laptop with AMD Ryzen 5000H processors and 32GB RAM for algorithm computation.
The TurtleBot4 communicates with the laptop via WiFi and ROS2 to transmit sensor information and control commands.
We set the dynamics limits as $v^r \in [0,0.26]$ m/s and $w^r \in [-\pi/3,\pi/3]$ rad/s.

In the first scenario, we conducted experiments in a classroom and the outside aisle.
Initially, the robot was in the aisle and searched for the target inside the classroom.
The initial belief distribution is characterized by a GMM consisting of four components with widely separated means and large covariance, creating dispersed particles to provide vague prior target information.
The first row of \cref{Fig:exp_process} displays the SAT process in the classroom scenario.
Specifically, despite significant prior uncertainty, the robot successfully completed the challenging target search task guided by critical particles from the hierarchical particle structure, and stably tracked the target with high visibility once the target was detected.

In the second scenario, we tested our approach in a cluttered space with various obstacles.
As shown in the second row of \Cref{Fig:exp_process}, the robot was capable of searching for the target under inaccurate target estimation, and can perform real-time and robust target tracking in narrow free space, even when the target rapidly turned between obstacles. 
Experiments in this scenario demonstrate the effectiveness of the proposed approach in environments with high obstacle density.

Furthermore, we record the target search time (ST), target visible rate (VR) during the tracking process, and overall algorithm computation frequency (CF) across three repeated trials in \Cref{tab:visibility_frequency}. 
The quantitative results show that our method achieves less search time and higher target visibility, demonstrating ReSPIRe's superior search efficiency and stable tracking of dynamic targets in various complex environments. 
Additionally, our approach achieves a computation frequency of 20-30 Hz during experiments, which is sufficient for online planning to address unknown environments and uncertainties while completing the SAT task. 
Similar to the simulation results, ReSPIRe consistently outperforms the optimal baseline in real-world experiments.

\subsection{Outdoor Experiments}
We used the Unitree Go2 to search for and track a walking person carried with an Apriltag in outdoor experiments.
The Unitree Go2 is equipped with an L1 Lidar for 2d mapping, an onboard RGB camera for visual detection, and a Jetson Orin Nano with 6-core ARM CPU and 8GB RAM for onboard computation. 
We deployed our algorithm on the onboard processor to complete SAT tasks. 
The dynamics limits are set as $v^r \in [0,1.2]$ m/s and $w^r \in [-\pi/3,\pi/3]$ rad/s.

\begin{figure}[t]
\centering
\includegraphics[width=\linewidth]{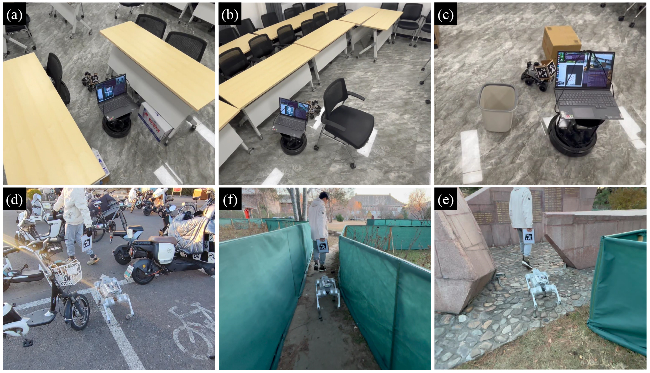}
\caption{\textbf{Target tracking in challenging environments.}
Challenging target tracking (a)-(b) in crowded desks and chairs, (c)-(d) among cluttered obstacles, (e) in a narrow walkway, (f) on different terrains.}
\label{Fig:exp_tracking_in_cluttered_env}
\end{figure}

In the first outdoor scenario, we tested our algorithm in a crowded bicycle parking lot.
As shown in the third row of~\Cref{Fig:exp_process}, the robot can rapidly detect the target and maintain robust tracking even in extremely cluttered environments.
We also conducted experiments in a park as the second outdoor scenario shown in the last row of~\Cref{Fig:exp_process}.
Despite the irregular terrain and diverse obstacles present in the extensive area, the robot still successfully accomplished complex SAT tasks.
\Cref{tab:visibility_frequency} also quantitatively validates the superiority of ReSPIRe in both search efficiency,  tracking performance, and computational efficiency compared to the optimal simulated baseline in outdoor experiments.

\begin{table}[t]
    \centering
    \caption{Real-world performance comparison across different scenarios}
    \small
    \begin{tabular}{cccc}
        \toprule
        Scenario & Metric & \multicolumn{1}{c}{PFT-50} & \multicolumn{1}{c}{ReSPIRe} \\
        \midrule
        \multirow{3}{*}{Classroom} 
          & ST (s) $\downarrow$ & $76.6$ & $\textbf{57.7}$ \\
          & VR (\%) $\uparrow$ & $91.1$ & $\textbf{98.4}$ \\
          & CF (Hz) $\uparrow$& $17.3$ & $\textbf{26.8}$ \\
        \midrule
        
        \multirow{3}{*}{Cluttered space}
          & ST (s) & $29.4$ & $\textbf{23.0}$ \\
          & VR (\%) & $90.5$ & $\textbf{98.6}$ \\
          & CF (Hz) & $18.6$ & $\textbf{30.2}$ \\
        \midrule
        
        \multirow{3}{*}{Parking lot}
          & ST (s) & $84.1$ & $\textbf{32.3}$ \\
          & VR (\%) & $85.3$ & $\textbf{95.5}$ \\
          & CF (Hz) & $14.8$ & $\textbf{24.2}$ \\
        \midrule
        
        \multirow{3}{*}{Park}
          & ST (s) & $106.3$ & $\textbf{24.0}$ \\
          & VR (\%) & $87.7$ & $\textbf{94.1}$ \\
          & CF (Hz) & $13.2$ & $\textbf{22.6}$ \\
        \bottomrule
    \end{tabular}
    \begin{tablenotes}
      \small
      \item ST = Search Time,VR = visible rate
    \\ CF = computational frequency
    \end{tablenotes}
    \label{tab:visibility_frequency}
\end{table}

Furthermore, we depict several challenging tracking conditions in cluttered environments shown in \Cref{Fig:exp_tracking_in_cluttered_env}, demonstrating ReSPIRe is competent for challenging tracking tasks in different scenarios and platforms.
The indoor and outdoor experiments in diverse environments demonstrate the planning efficiency and real-time operation capability of the proposed method in realistic complex scenarios characterized by unknown cluttered environments and imprecise target information with significant uncertainty.
The real-world experiments conducted on two robotic platforms also show the generalizability of the proposed approach to different robot platforms. 

\section{Conclusion}
This work has proposed ReSPIRe, an online, informative trajectory planner for mobile target SAT in unknown cluttered environments with considerably inaccurate prior target information.
We present a novel SP-based approximation approach to compute MI fast and accurately under non-Gaussian belief and continuous observation space.
A hierarchical particle structure has been introduced to extract critical information under dispersed particles, which flexibly adjusts the particle number for computational efficiency improvement.
To plan effectively and efficiently under uncertainty, the robot incrementally builds a policy tree by RBTS, which leverages previous rollout information to expand multiple nodes simultaneously, speeding up the tree construction process to improve planning efficiency.
We have conducted comprehensive simulations and real-world experiments to demonstrate the effectiveness of the proposed approach, revealing higher search efficiency, stable tracking performance, and real-time operational capability.
Future work will focus on SAT based on multi-robot coordination, and extend the proposed methods to more intricate scenarios.

\section*{Acknowledgement}
We thank Yi Zeng, Haoyang Song, Ieng Hou U, Yunze Hu for their help in experiments.

{
\bibliographystyle{ieeetr}
\bibliography{ref}

@string {ICRA = "ICRA"}

@string {RA-L = "RA-L"}

@string {ACC  = "ACC"}

@string {IJRR = "IJRR"}

@string {TRO = "T-RO"}

@string {TMECH="TMECH"}

@string {ICRA = "IEEE International Conference on Robotics and Automation"}

@string {ACC  = "American Control Conference"}

@string {RA-L = "IEEE Robotics and Automation Letters"}

@string {IJRR = "International Journal of Robotics Research"}

@string {TRO = "IEEE Transactions on Robotics"}

@string {TMECH = "IEEE/ASME Transactions on Mechatronics"}

@article{hoffmann2009mobile,
  title={Mobile sensor network control using mutual information methods and particle filters},
  author={Hoffmann, Gabriel M and Tomlin, Claire J},
  journal={IEEE Transactions on Automatic Control},
  volume={55},
  number={1},
  pages={32--47},
  year={2009},
  publisher={IEEE},
  note={doi:10.1109/tac.2009.2034206.}
}

@article{charrow2014approximate,
  title={Approximate representations for multi-robot control policies that maximize mutual information},
  author={Charrow, Benjamin and Kumar, Vijay and Michael, Nathan},
  journal={Autonomous Robots},
  volume={37},
  pages={383--400},
  year={2014},
  publisher={Springer},
  note={doi:10.15607/rss.2013.ix.053.}
}

@article{ghaffari2019sampling,
  title={Sampling-based incremental information gathering with applications to robotic exploration and environmental monitoring},
  author={Ghaffari Jadidi, Maani and Valls Miro, Jaime and Dissanayake, Gamini},
  journal=IJRR,
  volume={38},
  number={6},
  pages={658--685},
  year={2019},
  publisher={SAGE Publications Sage UK: London, England},
  note={doi:10.1177/0278364919844575.}
}

@article{julian2012distributed,
  title={Distributed robotic sensor networks: An information-theoretic approach},
  author={Julian, Brian J and Angermann, Michael and Schwager, Mac and Rus, Daniela},
  journal=IJRR,
  volume={31},
  number={10},
  pages={1134--1154},
  year={2012},
  publisher={SAGE Publications Sage UK: London, England},
  note={doi:10.1177/0278364912452675.}
}

@article{tisdale2009autonomous,
  title={Autonomous UAV path planning and estimation},
  author={Tisdale, John and Kim, ZuWhan and Hedrick, J Karl},
  journal={IEEE Robotics \& Automation Magazine},
  volume={16},
  number={2},
  pages={35--42},
  year={2009},
  publisher={IEEE},
  note={doi:10.1109/mra.2009.932529.}
}

@article{ryan2010particle,
  title={Particle filter based information-theoretic active sensing},
  author={Ryan, Allison and Hedrick, J Karl},
  journal={Robotics and Autonomous Systems},
  volume={58},
  number={5},
  pages={574--584},
  year={2010},
  publisher={Elsevier},
  note={doi:10.1016/j.robot.2010.01.001.}
}

@inproceedings{huber2008entropy,
  title={On entropy approximation for Gaussian mixture random vectors},
  author={Huber, Marco F and Bailey, Tim and Durrant-Whyte, Hugh and Hanebeck, Uwe D},
  booktitle={IEEE International Conference on Multisensor Fusion and Integration for Intelligent Systems},
  year={2008},
  note={doi:10.1109/mfi.2008.4648062.}
}

@inproceedings{liu2017model,
  title={Model predictive control-based target search and tracking using autonomous mobile robot with limited sensing domain},
  author={Liu, Chang and Hedrick, J Karl},
  booktitle=ACC,
  year={2017},
  note={doi:10.23919/acc.2017.7963397.}
}

@inproceedings{sunberg2018online,
  title={Online algorithms for POMDPs with continuous state, action, and observation spaces},
  author={Sunberg, Zachary and Kochenderfer, Mykel},
  booktitle={International Conference on Automated Planning and Scheduling},
  year={2018},
  note={doi:10.1609/icaps.v28i1.13882.}
}

@article{silver2010monte,
  title={Monte-Carlo planning in large POMDPs},
  author={Silver, David and Veness, Joel},
  journal={Advances in neural information processing systems},
  year={2010},
}

@article{somani2013despot,
  title={DESPOT: Online POMDP planning with regularization},
  author={Somani, Adhiraj and Ye, Nan and Hsu, David and Lee, Wee Sun},
  journal={Advances in neural information processing systems},
  year={2013},
  note={doi:10.1613/jair.5328.}
}

@inproceedings{kurniawati2016online,
  title={An online POMDP solver for uncertainty planning in dynamic environment},
  author={Kurniawati, Hanna and Yadav, Vinay},
  booktitle={Robotics Research: The 16th International Symposium ISRR},
  year={2016},
  organization={Springer},
  note={doi:10.1007/978-3-319-28872-7\_35.}
}

@article{best2019dec,
  title={Dec-MCTS: Decentralized planning for multi-robot active perception},
  author={Best, Graeme and Cliff, Oliver M and Patten, Timothy and Mettu, Ramgopal R and Fitch, Robert},
  journal=IJRR,
  volume={38},
  number={2-3},
  pages={316--337},
  year={2019},
  publisher={SAGE Publications Sage UK: London, England},
  note={doi:10.1177/0278364918755924.}
}

@article{goldhoorn2018searching,
  title={Searching and tracking people with cooperative mobile robots},
  author={Goldhoorn, Alex and Garrell, Ana{\'\i}s and Alqu{\'e}zar, Ren{\'e} and Sanfeliu, Alberto},
  journal={Autonomous Robots},
  volume={42},
  number={4},
  pages={739--759},
  year={2018},
  publisher={Springer},
  note={doi:10.1007/s10514-017-9681-6.}
}

@inproceedings{wandzel2019multi,
  title={Multi-object search using object-oriented pomdps},
  author={Wandzel, Arthur and Oh, Yoonseon and Fishman, Michael and Kumar, Nishanth and Wong, Lawson LS and Tellex, Stefanie},
  booktitle=ICRA,
  year={2019},
  note={doi:10.1109/icra.2019.8793888.}
}

@inproceedings{xiao2019online,
  title={Online planning for target object search in clutter under partial observability},
  author={Xiao, Yuchen and Katt, Sammie and ten Pas, Andreas and Chen, Shengjian and Amato, Christopher},
  booktitle=ICRA,
  year={2019},
  note={doi:10.1109/icra.2019.8793494.}
}

@inproceedings{furukawa2006recursive,
  title={Recursive Bayesian search-and-tracking using coordinated UAVs for lost targets},
  author={Furukawa, Tomonari and Bourgault, Frederic and Lavis, Benjamin and Durrant-Whyte, Hugh F},
  booktitle=ICRA,
  year={2006},
  note={doi:10.1109/robot.2006.1642081.}
}

@article{wolek2020cooperative,
  title={Cooperative mapping and target search over an unknown occupancy graph using mutual information},
  author={Wolek, Artur and Cheng, Sheng and Goswami, Debdipta and Paley, Derek A},
  journal=RA-L,
  volume={5},
  number={2},
  pages={1071--1078},
  year={2020},
  publisher={IEEE},
  note={doi:10.1109/lra.2020.2966394.}
}

@article{lozano2022surveillance,
  title={Surveillance and collision-free tracking of an aggressive evader with an actuated sensor pursuer},
  author={Lozano, Eliezer and Ruiz, Ubaldo and Becerra, Israel and Murrieta-Cid, Rafael},
  journal=RA-L,
  volume={7},
  number={3},
  pages={6854--6861},
  year={2022},
  publisher={IEEE},
  note={doi:10.1109/lra.2022.3178799.}
}

@inproceedings{fischer2020information,
  title={Information particle filter tree: An online algorithm for pomdps with belief-based rewards on continuous domains},
  author={Fischer, Johannes and Tas, {\"O}mer Sahin},
  booktitle={International Conference on Machine Learning},
  year={2020},
}

@inproceedings{kocsis2006bandit,
  title={Bandit based monte-carlo planning},
  author={Kocsis, Levente and Szepesv{\'a}ri, Csaba},
  booktitle={European conference on machine learning},
  year={2006},
  organization={Springer},
  note={doi:10.1007/11871842\_29.}
}

@book{van2004sigma,
  title={Sigma-point Kalman filters for probabilistic inference in dynamic state-space models},
  author={Van Der Merwe, Rudolph},
  year={2004},
  publisher={Oregon Health \& Science University},
}

@article{legrand2022cell,
  title={Cell multi-Bernoulli (cell-MB) sensor control for multi-object search-while-tracking (SWT)},
  author={LeGrand, Keith A and Zhu, Pingping and Ferrari, Silvia},
  journal={IEEE Transactions on Pattern Analysis and Machine Intelligence},
  volume={45},
  number={6},
  pages={7195--7207},
  year={2022},
  publisher={IEEE},
  note={doi:10.1109/tpami.2022.3223856.}
}

@article{sung2021gm,
  title={Gm-phd filter for searching and tracking an unknown number of targets with a mobile sensor with limited fov},
  author={Sung, Yoonchang and Tokekar, Pratap},
  journal={IEEE Transactions on Automation Science and Engineering},
  volume={19},
  number={3},
  pages={2122--2134},
  year={2021},
  publisher={IEEE},
  note={doi:10.31219/osf.io/hryg7.}
}

@article{kolchinsky2017estimating,
  title={Estimating mixture entropy with pairwise distances},
  author={Kolchinsky, Artemy and Tracey, Brendan D},
  journal={Entropy},
  volume={19},
  number={7},
  pages={361-377},
  year={2017},
  publisher={MDPI},
  note={doi:10.3390/e19070361.}
}

@article{zhang2020fsmi,
  title={FSMI: Fast computation of Shannon mutual information for information-theoretic mapping},
  author={Zhang, Zhengdong and Henderson, Theia and Karaman, Sertac and Sze, Vivienne},
  journal=IJRR,
  volume={39},
  number={9},
  pages={1155--1177},
  year={2020},
  publisher={SAGE Publications Sage UK: London, England},
  note={doi:10.1109/icra.2019.8793541.}
}

@inproceedings{zhou2024aspire,
  title={Aspire: An informative trajectory planner with mutual information approximation for target search and tracking},
  author={Zhou, Kangjie and Wu, Pengying and Su, Yao and Gao, Han and Ma, Ji and Liu, Hangxin and Liu, Chang},
  booktitle={2024 IEEE International Conference on Robotics and Automation (ICRA)},
  pages={4626--4632},
  year={2024},
  organization={IEEE}
}

@article{schlotfeldt2018anytime,
  title={Anytime planning for decentralized multirobot active information gathering},
  author={Schlotfeldt, Brent and Thakur, Dinesh and Atanasov, Nikolay and Kumar, Vijay and Pappas, George J},
  journal=RA-L,
  volume={3},
  number={2},
  pages={1025--1032},
  year={2018},
  publisher={IEEE},
  note={doi:10.1109/lra.2018.2794608.}
}

@article{nair2006entropy,
  title={On entropy for mixtures of discrete and continuous variables},
  author={Nair, Chandra and Prabhakar, Balaji and Shah, Devavrat},
  journal={arXiv preprint cs/0607075},
  year={2006},
}

@article{schmid2020efficient,
  title={An efficient sampling-based method for online informative path planning in unknown environments},
  author={Schmid, Lukas and Pantic, Michael and Khanna, Raghav and Ott, Lionel and Siegwart, Roland and Nieto, Juan},
  journal=RA-L,
  volume={5},
  number={2},
  pages={1500--1507},
  year={2020},
  publisher={IEEE},
  note={doi:10.1109/lra.2020.2969191.}
}

@article{asgharivaskasi2023semantic,
  title={Semantic octree mapping and shannon mutual information computation for robot exploration},
  author={Asgharivaskasi, Arash and Atanasov, Nikolay},
  journal=TRO,
  year={2023},
  volume={39},
  number={3},
  pages={1910-1928},
  publisher={IEEE},
  note={doi:10.1109/tro.2023.3245986.}
}

@article{tao2021path,
  title={Path planning in uncertain environment with moving obstacles using warm start cross entropy},
  author={Tao, Xiuye and Lang, Ning and Li, Huiping and Xu, Demin},
  journal=TMECH,
  volume={27},
  number={2},
  pages={800--810},
  year={2021},
  publisher={IEEE},
  note={doi:10.1109/tmech.2021.3071723.}
}

@article{wang2023active,
  title={Active view planning for visual slam in outdoor environments based on continuous information modeling},
  author={Wang, Zhihao and Chen, Haoyao and Zhang, Shiwu and Lou, Yunjiang},
  journal=TMECH,
  volume={29},
  number={1},
  pages={237--248},
  year={2023},
  publisher={IEEE},
  note={doi:10.1109/tmech.2023.3272910/mm1.}
}

@article{chen2023flying,
  title={Flying in Dynamic Scenes with Multitarget Velocimetry and Perception-Enhanced Planning},
  author={Chen, Han and Wen, Chih-Yung and Gao, Fei and Lu, Peng},
  journal=TMECH,
  volume={29},
  number={1},
  pages={521--532},
  year={2023},
  publisher={IEEE},
  note={doi:10.1109/tmech.2023.3289180.}
}

@article{xiao2017sampling,
  title={A sampling-based Bayesian approach for cooperative multiagent online search with resource constraints},
  author={Xiao, Hu and Cui, Rongxin and Xu, Demin},
  journal={IEEE Transactions on Cybernetics},
  volume={48},
  number={6},
  pages={1773--1785},
  year={2017},
  publisher={IEEE},
  note={doi:10.1109/tcyb.2017.2715228.}
}

@article{wang2019semantic,
  title={Semantic-aware informative path planning for efficient object search using mobile robot},
  author={Wang, Chaoqun and Cheng, Jiyu and Chi, Wenzheng and Yan, Tingfang and Meng, Max Q-H},
  journal={IEEE Transactions on Systems, Man, and Cybernetics: Systems},
  volume={51},
  number={8},
  pages={5230--5243},
  year={2019},
  publisher={IEEE},
  note={doi:10.1109/tsmc.2019.2946646.}
}

@article{papaioannou2023distributed,
  title={Distributed search planning in 3-d environments with a dynamically varying number of agents},
  author={Papaioannou, Savvas and Kolios, Panayiotis and Theocharides, Theocharis and Panayiotou, Christos G and Polycarpou, Marios M},
  journal={IEEE Transactions on Systems, Man, and Cybernetics: Systems},
  volume={53},
  number={7},
  pages={4117--4130},
  year={2023},
  publisher={IEEE},
  note={doi:10.1109/tsmc.2023.3240023.}
}

@article{cui2015mutual,
  title={Mutual information-based multi-AUV path planning for scalar field sampling using multidimensional RRT},
  author={Cui, Rongxin and Li, Yang and Yan, Weisheng},
  journal={IEEE Transactions on Systems, Man, and Cybernetics: Systems},
  volume={46},
  number={7},
  pages={993--1004},
  year={2015},
  publisher={IEEE},
  note={doi:10.1109/tsmc.2015.2500027.}
}

@article{li2024collision,
  title={Collision-free source seeking control methods for unicycle robots},
  author={Li, Tinghua and Jayawardhana, Bayu},
  journal={IEEE Transactions on Automatic Control},
  volume={70},
  number={3},
  pages={2020--2027},
  year={2024},
  publisher={IEEE},
  note={doi:10.1109/tsmc.2018.2808471.}
}

@article{kong2023collision,
  title={A collision-free target tracking controller with uncertain disturbance rejection for quadruped robots},
  author={Kong, Shihan and Sun, Jinlin and Luo, Aocheng and Chi, Wanchao and Zhang, Chong and Zhang, Shenghao and Liu, Yuzhen and Zhu, Qiuguo and Yu, Junzhi},
  journal={IEEE Transactions on Intelligent Vehicles},
  volume={9},
  number={1},
  pages={670--680},
  year={2023},
  publisher={IEEE},
  note={doi:10.1109/TIV.2023.3296669.}
}

@article{yu2025cognitive,
  title={Cognitive Robotics: Enhancing Multirobot Target Search in Unknown Environments Through Adaptive Communication Strategies},
  author={Yu, Xuewei and Su, Bo and Wang, Ziheng and Zhang, Jianlei and Zhang, Chunyan},
  journal={IEEE Transactions on Systems, Man, and Cybernetics: Systems},
  volume={55},
  number={5},
  pages={3449--3463},
  year={2025},
  publisher={IEEE},
  note={doi:10.1109/TSMC.2025.3540059.}
}

@article{senthilnath2024metacognitive,
  title={Metacognitive Decision-Making Framework for Multi-UAV Target Search Without Communication},
  author={Senthilnath, J and Harikumar, K and Sundaram, Suresh},
  journal={IEEE Transactions on Systems, Man, and Cybernetics: Systems},
  volume={54},
  number={5},
  pages={3195--3206},
  year={2024},
  publisher={IEEE},
  note={doi:10.1109/TSMC.2024.3358060.}
}

@article{cui2025cooperative,
  title={Cooperative Multi-AAV Path Planning for Discovering and Tracking Multiple Radio-Tagged Targets},
  author={Cui, Yukang and Chen, Jiabin and Lin, Hong and Shu, Zhan and Huang, Tingwen},
  journal={IEEE Transactions on Systems, Man, and Cybernetics: Systems},
  volume={55},
  number={4},
  pages={2463--2475},  
  year={2025},
  publisher={IEEE},
  note={doi:10.1109/TSMC.2024.3523132.}
}
}

\appendix
\subsection{Proof of \Cref{thm:definition}}
\label{appendix:reward_definition}
\begin{IEEEproof}
First, we calculate $H(\boldsymbol{z}_{k+1}|\boldsymbol{x}_{k+1}^t)$ with the particle expression,
\begin{equation}
\resizebox{0.89\linewidth}{!}{%
$\begin{aligned}
H(\boldsymbol{z}_{k+1}|\boldsymbol{x}_{k+1}^t) 
&\approx \sum\nolimits_{j=1}^{N} P(\tilde{\boldsymbol{x}}_{k+1}^{t,j})H(\boldsymbol{z}_{k+1}|\boldsymbol{x}_{k+1}^t=\tilde{\boldsymbol{x}}_{k+1}^{t,j})\\
&\approx \sum\nolimits_{j=1}^{N} w_{k}^{j}H(\boldsymbol{z}_{k+1}|\tilde{\boldsymbol{x}}_{k+1}^{t,j}),
\end{aligned}$}
\end{equation}
where the first equality is derived from the definition of conditional entropy and the second equality is obtained by substituting $P(\tilde{\boldsymbol{x}}_{k+1}^{t,j})$ with the particle-based future belief
\begin{equation}
\begin{split}
P(\boldsymbol{x}_{k+1}^t) \approx \sum\nolimits_{j=1}^{N}w_{k}^{j}\delta(\boldsymbol{x}_{k+1}^t-\tilde{\boldsymbol{x}}_{k+1}^{t,j}),
\end{split}
\end{equation} 
According to the observation model, the probability density function is
\begin{equation}
\resizebox{0.89\linewidth}{!}{%
$\begin{aligned}
P(\boldsymbol{z}_{k+1}|\tilde{\boldsymbol{x}}_{k+1}^{t,j}) 
=  \begin{cases}
\mathcal{N}(\boldsymbol{z}_{k+1};\mathbf{h}(\boldsymbol{x}_{k+1}^{r},\tilde{\boldsymbol{x}}_{k+1}^{t,j}),\mathbf{\Sigma}) & j \in S_f\\
{\mathds{1}}_{\boldsymbol{z}_{k+1}=\varnothing} & j \notin S_f
\end{cases},
\end{aligned}$}
\end{equation}
where $S_f$ denote the set of indices that particles are inside the FOV.
Note that the variable $\boldsymbol{z}_{k+1}$ is a continuous-discrete mixed variable defined in $\mathbb{R}^m \cup \varnothing$.
We first introduce the definition of entropy for mixed random variable~\cite{nair2006entropy}.
Given a mixed random variable $X$ that either takes discrete value $x_1$, $x_2$, $\ldots$ with probabilities $p_1$, $p_2$, $\ldots$ or follows a density function $f(x)$, satisfying the condition
\begin{equation}
\begin{split}
\int_{\mathbb{R}} \widetilde{f}(x) dx + \sum_i p_i = 1.
\end{split}
\end{equation}
where $\widetilde{f}(x) = (1-p)f(x)$ and $p = \sum_i p_i$.
The definition of the entropy for the mixed random variable $X$ is
\begin{equation}
\begin{split}
H(X)= -\sum_i p_i \log p_i - \int_{\mathbb{R}} \widetilde{f}(x)\log \widetilde{f}(x)dx.
\end{split}
\label{eqn:mixed_entropy_definition}
\end{equation}
Based on the above definition, we can derive that
\begin{equation}
\begin{split}
H(\boldsymbol{z}_{k+1}|\tilde{\boldsymbol{x}}_{k+1}^{t,j})= -p_{\varnothing}^j\log p_{\varnothing}^j - \int_{\mathbb{R}^m} p_{\mathbb{R}}^j\log p_{\mathbb{R}}^jd\boldsymbol{z}_{k+1}.
\end{split}
\end{equation}
where $p_{\varnothing}^j = P(\boldsymbol{z}_{k+1} = \varnothing|\boldsymbol{x}_{k+1}^{t,j})$ and $p_{\mathbb{R}}^j = (1-p_{\varnothing})P(\boldsymbol{z}_{k+1}\in \mathbb{R}_m|\boldsymbol{x}_{k+1}^{t,j})$.
Note that $p_{\varnothing}^j \log p_{\varnothing}^j $ is defined as $0$ when $p_{\varnothing}^j = 0$ in information theory.
If the $j$th particle can be detected, i.e., $j\in S_f$, then $p_{\varnothing}^j = 0$ and we can obtain
\begin{equation}
\resizebox{\linewidth}{!}{%
$\begin{aligned}
H(\boldsymbol{z}_{k+1}|\tilde{\boldsymbol{x}}_{k+1}^{t,j}) &= - \int_{\mathbb{R}^m} P(\boldsymbol{z}_{k+1}|\boldsymbol{x}_{k+1}^{t,j})\log P(\boldsymbol{z}_{k+1}|\boldsymbol{x}_{k+1}^{t,j})d\boldsymbol{z}_{k+1}\\
& = \frac{m}{2}(\log2\pi+1)+\frac{1}{2}\log|\mathbf{\Sigma}|.
\nonumber
\end{aligned}$}
\end{equation}
where the second equality is because $P(\boldsymbol{z}_{k+1}\in \mathbb{R}_m|\boldsymbol{x}_{k+1}^{t,j})$ is a $m$-dimensional Gaussian distribution and its entropy has explicit expression.
Otherwise, the $j$th particle is outside the FOV and $p_{\varnothing}^j = 1$, so the corresponding entropy is
\begin{equation}
\begin{split}
H(\boldsymbol{z}_{k+1}|\tilde{\boldsymbol{x}}_{k+1}^{t,j}) = -p_{\varnothing}\log p_{\varnothing} = 0.
\end{split}
\end{equation}
So we can derive that
\begin{equation}
\resizebox{0.89\linewidth}{!}{%
$\begin{aligned}
H(\boldsymbol{z}_{k+1}|\tilde{\boldsymbol{x}}_{k+1}^{t,j})={\mathds{1}}_{j\in S_f} \left[\frac{m}{2}(\log2\pi+1)+\frac{1}{2}\log|\mathbf{\Sigma}|\right].
\end{aligned}$}
\end{equation}

Next, we consider computing entropy $H(\boldsymbol{z}_{k+1})$. 
Utilizing the particle representation of target state distribution,
the observation distribution $P(\boldsymbol{z}_{k+1})$ can be computed as
\begin{equation}
\begin{split}
P(\boldsymbol{z}_{k+1}) & = \int P(\boldsymbol{z}_{k+1}|\boldsymbol{x}_{k+1}^t)P(\boldsymbol{x}_{k+1}^t)d\boldsymbol{x}_{k+1}^t \\
& \approx \sum\nolimits_{j=1}^{N}w_{k}^{j}P(\boldsymbol{z}_{k+1}|\tilde{\boldsymbol{x}}_{k+1}^{t,j}).
\end{split}
\end{equation}
Similarly, according to the definition of \cref{eqn:mixed_entropy_definition}, we can obtain
\begin{equation}
\begin{split}
H(\boldsymbol{z}_{k+1})= -p_{\varnothing}\log p_{\varnothing} - \int_{\mathbb{R}^m} p_{r}\log p_{r}d\boldsymbol{z}_{k+1}.
\end{split}
\end{equation}
where 
\begin{equation}
\begin{split}
p_{\varnothing} = P(\boldsymbol{z}_{k+1} = \varnothing) = \sum\nolimits_{j\notin S_f} w_{k}^{j},
\end{split}
\end{equation}
\begin{equation}
\begin{split}
p_{r} = \sum\nolimits_{j\in S_f} w_{k}^{j} P(\boldsymbol{z}_{k+1}|\tilde{\boldsymbol{x}}_{k+1}^{t,j}).
\end{split}
\end{equation}

\end{IEEEproof}

\subsection{Proof of \Cref{thm:sp}}
\label{appendix:sp}
We first explain how to obtain $\hat{H_r}$ to approximate $H_r$ with sigma points.
\begin{IEEEproof}
(SP-based entropy approximation method.)
From the definition, $H_r$ can be computed as
\begin{equation}
\begin{split}
H_r
&=-\int_{\mathbb{R}^{m}} p_r\log p_r d \boldsymbol{z}_{k+1} \approx -\sum\nolimits_{j\in S_f} w_{k}^{j} H_j,
\end{split}
\label{eqn:appendix_H_r}
\end{equation}
where 
\begin{equation}
\begin{split}
H_j = \int_{\mathbb{R}^{m}} P(\boldsymbol{z}_{k+1}|\tilde{\boldsymbol{x}}_{k+1}^{t,j})\log p_r d\boldsymbol{z}_{k+1}.
\end{split}
\end{equation}
Based on sigma points, for all $j \in S_f$, the $j$th Gaussian component can be approximated as
\begin{equation}
\begin{split}
P(\boldsymbol{z}_{k+1}|\tilde{\boldsymbol{x}}_{k+1}^{t,j})\approx\sum\nolimits_{l=0}^{2m}w_{s}^{j,l}\delta(\boldsymbol{z}_{k+1}-\tilde{\boldsymbol{z}}_{k+1}^{j,l}),
\end{split}
\end{equation}
and $H_j$ can be approximated as
\begin{equation} \label{eqn:sp_inFOV}
\begin{split}
H_j
& \approx \sum\nolimits_{l=0}^{2m}w_{s}^{j,l}\log p_r(\tilde{\boldsymbol{z}}_{k+1}^{j,l})\\
& = \sum\nolimits_{l=0}^{2m}w_{s}^{j,l}\log\sum\nolimits_{i\in S_f}  w_{k}^{i}P(\tilde{\boldsymbol{z}}_{k+1}^{j,l}|\tilde{\boldsymbol{x}}_{k+1}^{t,i}).
\end{split}
\end{equation}
Combined with \cref{eqn:appendix_H_r} and \cref{eqn:sp_inFOV}, we obtain an approximation of $H_r$ by
\begin{equation}
\resizebox{0.89\linewidth}{!}{%
$\begin{aligned}
\hat{H_r} = - \sum\nolimits_{j\in S_f} w_{k}^{j}\sum\nolimits_{l=0}^{2m}w_{s}^{j,l}\log\sum\nolimits_{i\in S_f}  w_{k}^{i}P(\tilde{\boldsymbol{z}}_{k+1}^{j,l}|\tilde{\boldsymbol{x}}_{k+1}^{t,i}).
\end{aligned}$}
\label{eqn:hat_H_r}
\end{equation}
\end{IEEEproof}


Next, we prove the approximation bound of the SP-based method.
In the following derivation, we omit the subscript representing the time for notational simplicity. 
To prove this theorem, we first need the following lemmas.


\begin{lemma}\label{lemma1}
(Hessian matrix of log-GMM.)
Let $p(\boldsymbol{z})=\sum_{i=1}^{N} w^i p_i(\boldsymbol{z})$ be a GMM where $p_i(\boldsymbol{z})=\mathcal{N}(\boldsymbol{z};\mu_i,\mathbf{\Sigma})$. Denote $g(\boldsymbol{z})=\log p(\boldsymbol{z})$. The Hessian matrix of $g$ is
\begin{equation}
\resizebox{0.89\linewidth}{!}{%
$\begin{aligned}
\mathbf{H}_g(\boldsymbol{z})= p^{-2}\sum_{i=1}^N\sum_{j=1}^N \chi_i \chi_j \mathbf{\Sigma}^{-1} (\mu_j - \mu_i) \mu_j^T \mathbf{\Sigma}^{-1} - \mathbf{\Sigma}^{-1}.
\end{aligned}$}
\end{equation}
where $\chi_i=w^i p_i(\boldsymbol{z})$.
\end{lemma} 

\begin{IEEEproof}
\label{proof1}
According to the definition of $p$, we have
\begin{equation}
\begin{split}
\nabla p(\boldsymbol{z}) = -\sum_{i=1}^N w^{i} p_i(\boldsymbol{z}) \cdot \mathbf{\Sigma}^{-1} (\boldsymbol{z}-\mu_i),
\end{split}
\end{equation}
\begin{equation}
\resizebox{0.89\linewidth}{!}{%
$\mathbf{H}_p(\boldsymbol{z}) = \sum_{i=1}^N w^{i} p_i(\boldsymbol{z}) \cdot  \big(\mathbf{\Sigma}^{-1} (\boldsymbol{z}-\mu_i)(\boldsymbol{z}-\mu_i)^T \mathbf{\Sigma}^{-1}-\mathbf{\Sigma}^{-1} \big).$}
\end{equation}
By the chain rule, we can obtain
\begin{equation}
\begin{split}
\mathbf{H}_g(\boldsymbol{z}) = -{p(\boldsymbol{z})}^{-2}\nabla p(\boldsymbol{z})\nabla p(\boldsymbol{z})^T+{p(\boldsymbol{z})}^{-1}\mathbf{H}_p(\boldsymbol{z}).
\end{split}
\end{equation}
Denote $\chi_i=w^{i}p_i(\boldsymbol{z})$ and $\phi_i=\mathbf{\Sigma}^{-1} (\boldsymbol{z}-\mu_i)$, and we arrive at the following expression:
\begin{equation}
\resizebox{0.89\linewidth}{!}{%
$\begin{aligned}
&\mathbf{H}_g(\boldsymbol{z}) \\
&= -{p(\boldsymbol{z})}^{-2}\nabla p(\boldsymbol{z})\nabla p(\boldsymbol{z})^T+{p(\boldsymbol{z})}^{-1}\mathbf{H}_p(\boldsymbol{z})\\
&= -p^{-2}\sum_{i=1}^N \chi_i \phi_i\sum_{j=1}^N \chi_j \phi_j^T + p^{-1} \sum_{i=1}^N \chi_i \phi_i\phi_i^T - \mathbf{\Sigma}^{-1}\\
&= p^{-2}\bigg( \sum_{i=1}^N \chi_i \sum_{j=1}^N \chi_j \phi_j\phi_j^T-\sum_{i=1}^N \chi_i \phi_i\sum_{j=1}^N \chi_j \phi_j^T\bigg) - \mathbf{\Sigma}^{-1}\\
&= p^{-2}\bigg(\sum_{i=1}^N\sum_{j=1}^N \chi_i \chi_j \phi_j \phi_j^T-\sum_{i=1}^N\sum_{j=1}^N \chi_i \chi_j \phi_i \phi_j^T \bigg) - \mathbf{\Sigma}^{-1}\\
&= p^{-2}\sum_{i=1}^N\sum_{j=1}^N \chi_i \chi_j (\phi_j-\phi_i) \phi_j^T - \mathbf{\Sigma}^{-1}\\
&= p^{-2}\sum_{i=1}^N\sum_{j=1}^N \chi_i \chi_j \mathbf{\Sigma}^{-1} (\mu_i - \mu_j) (\boldsymbol{z} - \mu_j)^T \mathbf{\Sigma}^{-1} - \mathbf{\Sigma}^{-1}\\
&= p^{-2}\sum_{i=1}^N\sum_{j=1}^N \chi_i \chi_j \mathbf{\Sigma}^{-1} (\mu_j - \mu_i) \mu_j^T \mathbf{\Sigma}^{-1} - \mathbf{\Sigma}^{-1}.
\end{aligned}$}
\end{equation}
The last equality is due to
\begin{equation}
\begin{split}
&\sum_{i=1}^N\sum_{j=1}^N \chi_i \chi_j \mathbf{\Sigma}^{-1} (\mu_i - \mu_j) (\boldsymbol{z} - \mu_j)^T \mathbf{\Sigma}^{-1}\\
&=  \mathbf{\Sigma}^{-1} \bigg(\sum_{i=1}^N\sum_{j=1}^N \chi_i \chi_j (\mu_i - \mu_j) (\boldsymbol{z} - \mu_j)^T \bigg) \mathbf{\Sigma}^{-1}\\
&=  \mathbf{\Sigma}^{-1}\bigg(\sum_{i=1}^N\sum_{j=1}^N \chi_i \chi_j  (\mu_i - \mu_j)\bigg) \boldsymbol{z}^T \mathbf{\Sigma}^{-1}\\& -\sum_{i=1}^N\sum_{j=1}^N \chi_i \chi_j \mathbf{\Sigma}^{-1} (\mu_i - \mu_j) \mu_j^T \mathbf{\Sigma}^{-1}\\
&=\sum_{i=1}^N\sum_{j=1}^N \chi_i \chi_j \mathbf{\Sigma}^{-1} (\mu_j - \mu_i) \mu_j^T \mathbf{\Sigma}^{-1}.
\end{split}
\end{equation} 
So the proof is completed.
\end{IEEEproof}

\begin{lemma}\label{lemma2}
$\forall \boldsymbol{x} \in \mathbb{R}^{m}$, the absolute value of the quadratic form of $\mathbf{G}$ is bounded with
\begin{equation}
\begin{split}
\big|\boldsymbol{x}^T \mathbf{G} \boldsymbol{x}\big|
&\leq  c \boldsymbol{x}^T\boldsymbol{x}.
\end{split}
\end{equation}
where
\begin{equation}
\begin{split}
\mathbf{G} 
&= p^{-2} \sum_{i=1}^N\sum_{j=1}^N \chi_i \chi_j \mathbf{\Sigma}^{-1} (\mu_j - \mu_i) \mu_j^T \mathbf{\Sigma}^{-1}.
\end{split}
\end{equation}
and $c$ is a relatively small positive constant.
\end{lemma}

\begin{IEEEproof}
According to the definition,
\begin{equation}
\begin{split}
\mathbf{G} 
&= p^{-2}\sum_{i=1}^N\sum_{j=1}^N \chi_i \chi_j \mathbf{\Sigma}^{-1} (\mu_j - \mu_i) \mu_j^T \mathbf{\Sigma}^{-1}\\
&= p^{-2}\sum_{i=1}^N\sum_{j>i}^N \chi_i \chi_j \mathbf{\Sigma}^{-1} (\mu_j - \mu_i) (\mu_j-\mu_i)^T \mathbf{\Sigma}^{-1}.\\
\end{split}
\end{equation}
Denote $\mathbf{v}_{ij} = \mathbf{\Sigma}^{-1} (\mu_j - \mu_i)$ and $\mathbf{G}_{ij}=\mathbf{v}_{ij} \mathbf{v}_{ij}^T$. With the particle filter, we can assume that
\begin{equation}
\begin{split}
{||\mathbf{v}_{ij}||}^2\leq c,
\end{split}
\end{equation}
where $c$ is a positive constant. Since $\mathbf{G}_{ij}$ is a rank one matrix, it only has eigenvalue $0$ and $\mathbf{v}_{ij}^T \mathbf{v}_{ij}$. So $\forall \boldsymbol{x} \in \mathbb{R}^{m}$, we have
\begin{equation}
\begin{split}
0 \leq \boldsymbol{x}^T \mathbf{G}_{ij} \boldsymbol{x}
\leq \mathbf{v}_{ij}^T \mathbf{v}_{ij}\boldsymbol{x}^T\boldsymbol{x}
\leq c\boldsymbol{x}^T\boldsymbol{x}
\end{split}
\end{equation}
and
\begin{equation}
\begin{split}
0 \leq \boldsymbol{x}^T \mathbf{G} \boldsymbol{x}
&= p^{-2}\sum_{i=1}^N\sum_{j>i}^N \chi_i \chi_j \boldsymbol{x}^T\mathbf{G}_{ij}\boldsymbol{x}\\
&\leq p^{-2}\sum_{i=1}^N\sum_{j>i}^N \chi_i \chi_j c \boldsymbol{x}^T\boldsymbol{x}\\
&\leq p^{-2}c \boldsymbol{x}^T\boldsymbol{x} \sum_{i=1}^N\sum_{j=1}^N \chi_i \chi_j \\
&= c \boldsymbol{x}^T\boldsymbol{x}.
\end{split}
\end{equation}
So we can derive
\begin{equation}
\begin{split}
\big|\boldsymbol{x}^T \mathbf{G} \boldsymbol{x}\big|
&\leq c \boldsymbol{x}^T\boldsymbol{x}.
\end{split}
\end{equation}
\end{IEEEproof}

\begin{lemma}\label{lemma3}
Let $\varphi_i = \big(\sqrt{(\lambda+m)\mathbf{\Sigma}}\big)_{i}$, then
\begin{equation}
\begin{split}
\varphi_i^T \varphi_i
&\leq  (\lambda+m)\sigma_{\max}^2,
\end{split}
\end{equation}
and
\begin{equation}
\begin{split}
\varphi_i^T \mathbf{\Sigma}^{-1} \varphi_i=  \lambda+m.
\end{split}
\end{equation}
\end{lemma}

\begin{IEEEproof}
Since covariance matrix $\mathbf{\Sigma}$ is a symmetric matrix, it can be diagonalized by $\mathbf{\Sigma}=\mathbf{U}^T\mathbf{\Lambda}\mathbf{U}$ with orthogonal matrix $\mathbf{U}$, where $\mathbf{\Lambda} = diag(\sigma_{1}^{2},\sigma_{2}^{2},\ldots,\sigma_{m}^{2})$. It is obvious that these diagonal elements are also the eigenvalues of $\mathbf{\Sigma}$.
We obtain
\begin{equation}
\begin{split}
\varphi_i &= \big(\sqrt{(\lambda+m)\mathbf{\Sigma}}\big)_{i}\\
&= \sqrt{(\lambda+m)}(\mathbf{\Sigma}^{\frac{1}{2}})_i\\
&= \sqrt{(\lambda+m)}(\mathbf{U}^T\mathbf{\Lambda}^{\frac{1}{2}}\mathbf{U})_{i}\\
&= \sqrt{(\lambda+m)}\mathbf{U}^T\mathbf{\Lambda}^{\frac{1}{2}}\mathbf{U}_{i},
\end{split}
\end{equation}
where $\mathbf{U}_{i}$ is the $i$th column of the orthogonal matrix $\mathbf{U}$. Then we have
\begin{equation}
\begin{split}
\varphi_i^T \varphi_i 
&= (\lambda+m)\mathbf{U}_{i}^T \mathbf{\Lambda}^{\frac{1}{2}}  \mathbf{U} \mathbf{U}^T \mathbf{\Lambda}^{\frac{1}{2}} \mathbf{U}_{i}\\
&= (\lambda+m)\mathbf{U}_{i}^T \mathbf{\Lambda} \mathbf{U}_{i}\\
&\leq (\lambda+m)\sigma_{\max}^2 \mathbf{U}_{i}^T\mathbf{U}_{i}\\
&= (\lambda+m)\sigma_{\max}^2,
\end{split}
\end{equation}
and
\begin{equation}
\begin{split}
\varphi_i^T \mathbf{\Sigma}^{-1}\varphi_i 
&= (\lambda+m)\mathbf{U}_{i}^T \mathbf{\Lambda}^{\frac{1}{2}}  \mathbf{U} \mathbf{\Sigma}^{-1}\mathbf{U}^T \mathbf{\Lambda}^{\frac{1}{2}} \mathbf{U}_{i}\\
&= (\lambda+m)\mathbf{U}_{i}^T \mathbf{\Lambda}^{\frac{1}{2}}\mathbf{\Lambda}^{-1}\mathbf{\Lambda}^{\frac{1}{2}} \mathbf{U}_{i}\\
&= (\lambda+m)\mathbf{U}_{i}^T\mathbf{U}_{i}\\
&= \lambda+m.
\end{split}
\end{equation}
\end{IEEEproof}

\begin{lemma}\label{lemma4}
Let $f(\boldsymbol{y}) = \mathcal{N}(\boldsymbol{y};\mathbf{0},\mathbf{\Sigma})$, then
\begin{equation}
\begin{split}
\int f(\boldsymbol{y}) \boldsymbol{y}^T \mathbf{\Sigma}^{-1} \boldsymbol{y} d\boldsymbol{y} = m.
\end{split}
\end{equation}
\end{lemma}

\begin{IEEEproof}
This essentially calculates the expectation of the quadric form $\mathbb{E} \left[ \boldsymbol{y}^{T}\mathbf{\Sigma}^{-1}\boldsymbol{y} \right]$, which is simple to compute by
\begin{equation}
\begin{split}
\int f(\boldsymbol{y}) \boldsymbol{y}^T \mathbf{\Sigma}^{-1} \boldsymbol{y} d\boldsymbol{y} 
&= \mathbb{E} \left[ \boldsymbol{y}^{T}\mathbf{\Sigma}^{-1}\boldsymbol{y} \right] \\
&= \text{tr} \left[ \mathbf{\Sigma}^{-1} \mathbf{\Sigma} \right] + \boldsymbol{0}^{T}\mathbf{\Sigma}^{-1}\boldsymbol{0} \\
&= \text{tr} \left[ I_m \right] \\
&= m.
\end{split}
\end{equation}
\end{IEEEproof}

Based on the above lemmas, we prove \Cref{thm:sp} as follows:

\begin{IEEEproof}
(Error bound of SP-based approximation method.)
According to \cref{eqn:appendix_H_r} and \cref{eqn:hat_H_r}, we denote $H_r = \sum\nolimits_{j\in S_f} w_{k}^{j} H_j$ and $\hat{H_r} = \sum\nolimits_{j\in S_f} w_{k}^{j} \hat{H_j}$, where
\begin{equation}
\begin{split}
H_j = -\int_{\mathbb{R}^{m}} p(\boldsymbol{z}|\tilde{\boldsymbol{x}}^{t,j}) g(\boldsymbol{z}) d\boldsymbol{z},
\end{split}
\end{equation}
\begin{equation}
\begin{split}
\hat{H_j} = -\sum\nolimits_{l=0}^{2m}w_{s}^{j,l}g(\tilde{\boldsymbol{z}}^{j,l}).
\end{split}
\end{equation}
According to Taylor's theorem, there exists variables $\xi_i \in (\mu_{j},\mu_{j}+\varphi_{i}), \eta_i \in (\mu_{j}-\varphi_{i},\mu_{j})$, such that
\begin{equation}
\begin{split}
g(\mu_{j}+\varphi_i) &= g(\mu_{j}) + \nabla g(\mu_{j})^T \varphi_i + \frac{1}{2} \varphi_i^T \mathbf{H}_g(\xi_{i}) \varphi_i\\
g(\mu_{j}-\varphi_i) &= g(\mu_{j}) - \nabla g(\mu_{j})^T \varphi_i + \frac{1}{2} \varphi_i^T \mathbf{H}_g(\eta_{i}) \varphi_i.
\end{split}
\end{equation}
Based on the definition of the sigma points (\cref{eqn:sigma_points}), $\hat{H_j}$ is computed as
\begin{equation}
\resizebox{0.89\linewidth}{!}{%
$\begin{aligned}
\hat{H_j} 
&=\sum_{l=0}^{2m}w_{s}^{j,l} g(\tilde{\boldsymbol{z}}^{j,l})\\
&=\frac{\lambda}{\lambda+m}g(\mu_{j})+\frac{1}{2(\lambda+m)}\sum_{i=1}^{m}\big(g(\mu_{j}+\varphi_i)+g(\mu_{j}-\varphi_i) \big)\\
&=g(\mu_{j})+\frac{1}{4(\lambda+m)}\sum_{i=1}^{m}\varphi_i^T\big(\mathbf{H}_g(\xi_{i})+\mathbf{H}_g(\eta_{i})\big)\varphi_i\\
&=g(\mu_{j})+\frac{1}{4(\lambda+m)}\sum_{i=1}^{m}\varphi_i^T\big(\mathbf{G} (\xi_{i})+\mathbf{G}(\eta_{i})\big)\varphi_i\\&-\frac{1}{2(\lambda+m)}\sum_{i=1}^{m}\varphi_i^T\mathbf{\Sigma}^{-1}\varphi_i\\
&=g(\mu_{j})+\frac{1}{4(\lambda+m)}\sum_{i=1}^{m}\varphi_i^T\big(\mathbf{G} (\xi_{i})+\mathbf{G}(\eta_{i})\big)\varphi_i -\frac{m}{2}.
\end{aligned}$}
\end{equation}
where the fourth equation is due to Lemma 1 and the last equation is due to Lemma 3.
Denote $\phi=\frac{1}{4(\lambda+m)}\sum_{i=1}^{m}\varphi_i^T\big(\mathbf{G} (\xi_{i})+\mathbf{G}(\eta_{i})\big)\varphi_i$ and the error is
\begin{equation}
\resizebox{0.89\linewidth}{!}{%
$\begin{aligned}
|H_j-\hat{H_j}|
&=\bigg|\int p(\boldsymbol{z}|\tilde{\boldsymbol{x}}^{t,j})g(\boldsymbol{z})d\boldsymbol{z}-(g(\mu_{j})+\phi-\frac{m}{2}) \bigg|\\
&=\bigg|\int p(\boldsymbol{z}|\tilde{\boldsymbol{x}}^{t,j})(g(\boldsymbol{z})-g(\mu_{j}))d\boldsymbol{z}-\phi +\frac{m}{2} \bigg|.
\end{aligned}$}
\label{eqn:error}
\end{equation}
Let $\boldsymbol{y}=\boldsymbol{z}-\mu_{j}$ and the Gaussian distribution $p(\boldsymbol{z}|\tilde{\boldsymbol{x}}^{t,j})$ is transformed as
\begin{equation}
\begin{split}
p(\boldsymbol{z}|\tilde{\boldsymbol{x}}^{t,j})
=\mathcal{N}(\boldsymbol{z};\mu_j,\mathbf{\Sigma})
=\mathcal{N}(\boldsymbol{y};\mathbf{0},\mathbf{\Sigma})
=p(\boldsymbol{y}|\tilde{\boldsymbol{x}}^{t,j}),
\end{split}
\end{equation}
and according to Taylor's theorem, there exists $\zeta_j$ between $\mu_{j}$ and $\boldsymbol{z}$ such that
\begin{equation}
\begin{split}
g(\boldsymbol{z})=g(\mu_{j})+\nabla g(\mu_{j})^T \boldsymbol{y} + \frac{1}{2} \boldsymbol{y}^T \mathbf{H}_g(\zeta_j) \boldsymbol{y},
\end{split}
\end{equation}
and the first quantity in \cref{eqn:error} can be computed as
\begin{equation}
\resizebox{0.89\linewidth}{!}{%
$\begin{aligned}
&\int p(\boldsymbol{z}|\tilde{\boldsymbol{x}}^{t,j})(g(\boldsymbol{z})-g(\mu_{j}))d\boldsymbol{z} \\
&= \int p(\boldsymbol{y}|\tilde{\boldsymbol{x}}^{t,j}) \frac{1}{2} \boldsymbol{y}^T \mathbf{H}_g(\zeta_j) \boldsymbol{y} d\boldsymbol{y}\\
&= \frac{1}{2}\int p(\boldsymbol{y}|\tilde{\boldsymbol{x}}^{t,j}) \boldsymbol{y}^T \mathbf{G}(\zeta_j) \boldsymbol{y} d\boldsymbol{y}- \frac{1}{2}\int p(\boldsymbol{y}|\tilde{\boldsymbol{x}}^{t,j}) \boldsymbol{y}^T \mathbf{\Sigma}^{-1} \boldsymbol{y} d\boldsymbol{y}\\
&= \frac{1}{2}\int p(\boldsymbol{y}|\tilde{\boldsymbol{x}}^{t,j}) \boldsymbol{y}^T \mathbf{G}(\zeta_j) \boldsymbol{y} d\boldsymbol{y}- \frac{m}{2}.
\end{aligned}$}
\end{equation}
where the last equation is based on Lemma 4.
So the error can be computed as
\begin{equation}
\begin{split}
|H_j-\hat{H_j}|
&=\bigg|\frac{1}{2}\int p(\boldsymbol{y}|\tilde{\boldsymbol{x}}^{t,j}) \boldsymbol{y}^T \mathbf{G}(\zeta_j) \boldsymbol{y} d\boldsymbol{y}-\phi \bigg|\\
&\leq \bigg|\frac{1}{2}\int p(\boldsymbol{y}|\tilde{\boldsymbol{x}}^{t,j}) \boldsymbol{y}^T \mathbf{G}(\zeta_j) \boldsymbol{y} d\boldsymbol{y}\bigg| + |\phi|\\
&\leq \frac{1}{2}\int p(\boldsymbol{y}|\tilde{\boldsymbol{x}}^{t,j}) \big|\boldsymbol{y}^T \mathbf{G}(\zeta_j) \boldsymbol{y} \big| d\boldsymbol{y}+ |\phi|\\
&\leq \frac{1}{2}c\int p(\boldsymbol{y}|\tilde{\boldsymbol{x}}^{t,j}) \boldsymbol{y}^T\boldsymbol{y} d\boldsymbol{y}+ |\phi|\\
&= \frac{1}{2}c \sum_{i=1}^{m} \mathbf{\Sigma}_{ii}+ |\phi|\\
&= \frac{1}{2}c \sum_{i=1}^{m} \sigma_{i}^2+ |\phi|.
\end{split}
\end{equation}
and
\begin{equation}
\begin{split}
|\phi|&=\frac{1}{4(\lambda+m)}\bigg|\sum_{i=1}^{m}\varphi_i^T\big(\mathbf{G} (\xi_{i})+\mathbf{G}(\eta_{i})\big)\varphi_i\bigg|\\
&\leq\frac{1}{4(\lambda+m)}\sum_{i=1}^{m}\big|\varphi_i^T\mathbf{G} (\xi_{i})\varphi\big|+\big|\varphi_i^T\mathbf{G} (\eta_{i})\varphi\big|\\
&\leq\frac{1}{4(\lambda+m)}\sum_{i=1}^{m}2c \varphi_i^T\varphi_i\\
&\leq\frac{c m\sigma_{\max}^2}{2}\\
\end{split}
\end{equation}
Then $|H_j-\hat{H_j}|$ can be bounded as
\begin{equation}
\begin{split}
|H_j-\hat{H_j}|	
&\leq \frac{1}{2}c \sum_{i=1}^{m} \sigma_{i}^2+\frac{c m\sigma_{\max}^2}{2}\\
&\leq c m\sigma_{\max}^2.
\end{split}
\end{equation}
Based on this, we can derive that
\begin{equation}
\begin{split}
|H_r-\hat{H_r}|	
& = \bigg|\sum\nolimits_{j\in S_f} w_{k}^{j} (H_j-\hat{H_j})\bigg|\\
&\leq \sum\nolimits_{j\in S_f} w_{k}^{j} |H_j-\hat{H_j}|	\\
&\leq c m\sigma_{\max}^2.
\end{split}
\end{equation}
\end{IEEEproof}






\end{document}